\documentclass[10pt,journal,cspaper,compsoc]{IEEEtran}

\usepackage{algorithm}
\usepackage{booktabs}
\usepackage{epsfig}
\usepackage{graphicx}
\usepackage{amsmath}
\usepackage{amssymb}
\usepackage{color} 
\usepackage{algorithmic}
\usepackage{subfigure}
\usepackage{url}
\usepackage{multirow}
\usepackage[usenames,dvipsnames,table]{xcolor}
 
\usepackage[dvipsnames]{xcolor}
\newcommand{\blue}[1]{\textcolor{black}{{#1}}}

\begin{document}

\title{PSGAN++: Robust Detail-Preserving Makeup Transfer and Removal}

\author{Si Liu, Wentao Jiang, Chen Gao, Ran He, Jiashi Feng, Bo Li, Shuicheng Yan,~\IEEEmembership{Fellow,~IEEE} \thanks{
Si Liu, Wentao Jiang, Chen Gao and Bo Li are with Institute of Artificial Intelligence, Beihang University, Email: \{liusi, jiangwentao, boli\}@buaa.edu.cn and gaochen.ai@gmail.com. Ran He is with Institute of Automation, Chinese Academy of Sciences, E-mail: ran.he@ia.ac.cn. Jiashi Feng is with National University of Singapore. Shuicheng Yan is with Sea AI Lab (SAIL). Ran He is the corresponding author.} }

\markboth{IEEE TRANSACTIONS ON PATTERN ANALYSIS AND MACHINE INTELLIGENCE, VOL. XX, NO. X, X 20XX}
{Shell \MakeLowercase{\textit{et al.}}: Bare Demo of IEEEtran.cls for Computer Society Journals}

\IEEEcompsoctitleabstractindextext{

\begin{abstract}
    In this paper, we address the makeup transfer and removal tasks simultaneously, which aim to transfer the makeup from a reference image to a source image and remove the makeup from the with-makeup image respectively.
    Existing methods have achieved much advancement in constrained scenarios, but it is still very challenging for them to transfer makeup between images with large pose and expression differences, or handle makeup details like blush on cheeks or highlight on the nose.
    In addition, they are hardly able to control the degree of makeup during transferring or to transfer a specified part in the input face.
    These defects limit the application of previous makeup transfer methods to real-world scenarios.
    In this work, we propose a Pose and expression robust Spatial-aware GAN (abbreviated as PSGAN++).
    PSGAN++ is capable of performing both detail-preserving makeup transfer and effective makeup removal.
    For makeup transfer, PSGAN++ uses a Makeup Distill Network (MDNet) to extract makeup information, which is embedded into spatial-aware makeup matrices.
    We also devise an Attentive Makeup Morphing (AMM) module that specifies how the makeup in the source image is morphed from the reference image, and a makeup detail loss to supervise the model within the selected makeup detail area.
    On the other hand, for makeup removal, PSGAN++ applies an Identity Distill Network (IDNet) to embed the identity information from with-makeup images into identity matrices.
    Finally, the obtained makeup/identity matrices are fed to a Style Transfer Network (STNet) that is able to edit the feature maps to achieve makeup transfer or removal.
    To evaluate the effectiveness of our PSGAN++, we collect a Makeup Transfer In the Wild (MT-Wild) dataset that contains images with diverse poses and expressions and a Makeup Transfer High-Resolution (MT-HR) dataset that contains high-resolution images.
    Experiments demonstrate that PSGAN++ not only achieves state-of-the-art results with fine makeup details even in cases of large pose/expression differences but also can perform partial or degree-controllable makeup transfer.
    Both the code and the newly collected datasets will be released at \url{https://github.com/wtjiang98/PSGAN}.
\end{abstract}

\begin{keywords}
Makeup Transfer, Makeup Removal, Generative Adversarial Networks.
\end{keywords}
}

\maketitle
\IEEEdisplaynotcompsoctitleabstractindextext
\IEEEpeerreviewmaketitle

\vspace{-6mm}
\section{Introduction}

Makeup transfer aims to transfer the makeup from an arbitrary reference face to a source face, which is widely demanded in many real-world scenarios.
For example, users of short video apps (e.g., Snapchat, Vine) or online chatting apps (like WhatsApp) would often like to beautify themselves in order to look better; on some e-commerce sites like Amazon, regarding cosmetic products especially,  the purchase intention of users would greatly increase after they virtually try these cosmetics and observe the effects of makeup on their faces.
On the contrary, makeup removal is to remove the makeup given a with-makeup image.
Makeup removal can be used in face verification tasks to improve accuracy.

\begin{table}[t]
    \centering
    \setlength{\tabcolsep}{1mm}{
        \begin{tabular}{@{}lccccc@{}}
            \toprule
            \multirow{2}{*}{Method}                                & \multicolumn{4}{c}{makeup transfer}                         & \multirow{2}{*}{makeup removal} \\ \cmidrule(lr){2-5}
                                                                   & robust   & partial      & degree          & detail     &                                 \\ \midrule
            BGAN \cite{Li2018BeautyGANIF}         &       &       &         &       &                          \\ \midrule
            PGAN \cite{Chang2018PairedCycleGANAS} &       &       &         &       & $\checkmark$                        \\ \midrule
            BGlow \cite{ChenBeautyGlow2019}       &       &       & $ \checkmark $ &       &                          \\ \midrule
            LADN \cite{Gu2019LADNLA}              &       &       & $ \checkmark $ &       & $\checkmark$                        \\ \midrule
            PSGAN \cite{jiang2020psgan}           & $\checkmark$ & $\checkmark$ & $ \checkmark $ &       &                          \\
            \midrule
            PSGAN++                                                & $\checkmark$ & $\checkmark$ & $\checkmark$   & $\checkmark$ & $\checkmark$                    \\ \bottomrule
            \end{tabular}
    }
    \caption{Comparison with existing methods.
    We use ``robust'' to denote pose/expression robustness, ``degree'' to denote degree-controllable transfer and ``detail'' to denote detail-preserving transfer. }
    \label{t1_compare_with_existing}
    \vspace{-11mm}
\end{table}

\begin{figure*}[!t]
    \centering
    \includegraphics[width=1.0\linewidth]{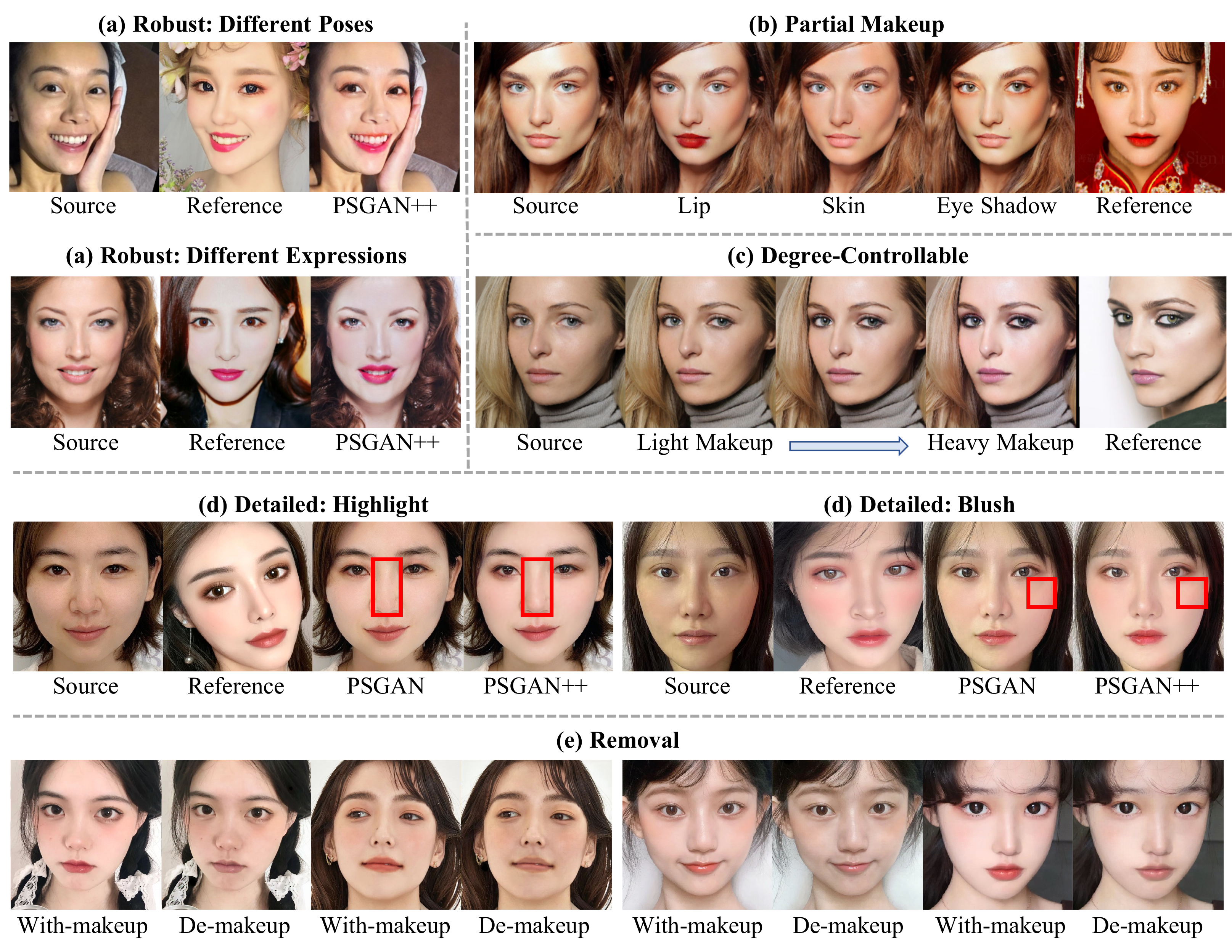}
    \caption{The makeup transfer results of PSGAN++ in terms of pose and expression robust, partial, degree-controllable, detail-preserving (highlights and blush) makeup transfer. Makeup removal results are also shown. }
    \vspace{-5mm}
    \label{fig:first_image}
\end{figure*}

In this work, we tackle joint makeup transfer and removal, targeting at real-world application requirements, and propose a novel method termed Pose and expression robust Spatial-aware GAN (abbreviated as PSGAN++).
The comparisons of our PSGAN++ with existing methods \cite{Li2018BeautyGANIF,ChenBeautyGlow2019,Chang2018PairedCycleGANAS,Gu2019LADNLA} are summarized in Table~\ref{t1_compare_with_existing}.
PSGAN++ is a fairly versatile method that is capable of performing robust, partial, makeup degree-controllable and detail-preserving makeup transfer as well as effective makeup removal within one framework.
($\romannumeral1$)
PSGAN++ can \emph{robustly} transfer the makeup between reference and source images even in case of very different poses and expressions, while existing methods are limited to frontal facial images with neutral expressions.
As shown in Figure~\ref{fig:first_image}(a), PSGAN++ produces high-quality results even though the poses and expressions are different between reference and source.
($\romannumeral2$)
PSGAN++ is able to perform \emph{partial} makeup transfer, i.e., transferring the makeup of specified facial regions separately, while existing methods cannot.
Some partial makeup results of PSGAN++ are illustrated in Figure~\ref{fig:first_image}(b), where the lip gloss, skin, and eye shadow can be individually transferred.
($\romannumeral3$)
PSGAN++ can produce \emph{makeup degree-controllable} makeup transfer results.
By makeup degree, here we mean the heaviness or lightness of makeup effects on human faces.
Figure \ref{fig:first_image}(c) shows the makeup degree of the transferred results is controllable from light to heavy with the proposed PSGAN++.
($\romannumeral4$)
PSGAN++ is the only method that can transfer \emph{makeup details}, including highlight on the nose and blush around cheeks, as shown in Figure \ref{fig:first_image}(d).
($\romannumeral5$)
PSGAN++ is also capable of achieving automatic \emph{makeup removal}, as shown in Figure \ref{fig:first_image}(e).

In structure, our PSGAN++ consists of a Makeup Distill Network (MDNet), an Attentive Makeup Morphing (AMM) module, a Style Transfer Network (STNet), and an Identity Distill Network (IDNet), to perform makeup transfer or removal through scaling or shifting the feature map for only once.
\emph{For makeup transfer}, the MDNet distills the makeup style from the reference image and embeds such feature information into two makeup matrices, which have the same spatial dimensions as visual features.
Then, these two makeup matrices are morphed and adapted to the source image by the AMM module to produce adapted makeup matrices.
Concretely, the AMM module utilizes the face parsing maps and facial landmarks to build pixel-wise correspondences between source and reference images, producing an attentive matrix to align faces.
Finally, the STNet achieves makeup transfer through applying pixel-wise multiplication and addition on visual features using the adapted makeup matrices.
To better transfer makeup details in the reference images, we also propose a makeup detail loss.
We first detect dense facial landmarks over both source images and reference images with a dense face alignment method \cite{feng2018prn}, and then select the landmarks inside the makeup detail areas (e.g., nose, cheek), based on which the makeup detail loss calculates the L1 difference between source and reference images in the selected areas.
In this way, PSGAN++ learns to generate the same makeup details as the reference.
\emph{For makeup removal}, the IDNet distills the identity information from the source image and embeds it into two identity matrices, and then the STNet achieves makeup removal by using these identity matrices to scale and shift the feature maps.
Through IDNet, PSGAN++ learns to automatically remove the makeup style, which may benefit other tasks such as face verification.

To evaluate our method over high-resolution images and images with diverse poses and expressions, we collect a Makeup Transfer In the Wild (MT-Wild) dataset that contains $772$ images with various poses and expressions and a Makeup Transfer High Resolution (MT-HR) dataset that contains $3,000$ images of $512 \times 512$ resolution.
To the best of our knowledge, MT-HR is the first makeup transfer dataset with $512 \times 512$ resolution.
Extensive results on three datasets, including Makeup Transfer (MT) \cite{Li2018BeautyGANIF}, the newly collected MT-Wild and MT-HR well demonstrate the effectiveness and superiority of the proposed PSGAN++.
Furthermore, we directly apply our method to facial videos in a frame-wise way and successfully attain nice and stable video makeup transfer results, which demonstrates the robustness and stability of our method.

We make the following contributions in this paper:
\begin{itemize}
   \item We develop PSGAN++, a versatile method that is able to perform pose/expression-robust, partial, makeup degree-controllable, and detail-preserving makeup transfer, as well as makeup removal.
   Our code will be released at \url{https://github.com/wtjiang98/PSGAN}.
   \item We propose MDNet/IDNet to extract makeup/identity information from the images to obtain spatial-aware makeup/identity matrices, respectively. The spatial-aware makeup matrices enable flexible partial and degree-controllable transfer while the identity matrices enable effective makeup removal.
   \item We devise an AMM module to adaptively morph makeup matrices to source images, enabling the pose and expression-robust transfer.
   Besides, a makeup detail loss is proposed for achieving makeup transfer with makeup details well preserved.
   \item We will release the newly built MT-Wild dataset that contains images with diverse poses and expressions, and the new MT-HR dataset with $512 \times 512$ images for facilitating further research on makeup transfer over high-resolution images.
\end{itemize}

This paper is an extension of our previous conference version \cite{jiang2020psgan}.
The current work adds to the initial version with significant novelties.
First, PSGAN++ proposes a new makeup detail loss to realize detail-preserving transfer including highlight and blush transfer.
Second, PSGAN++ can remove the makeup automatically by a newly proposed IDNet which is trained end-to-end with other sub-networks.
We also prove that makeup removal can improve the accuracy of face verification.
Third, we collect a new MT-HR dataset with $512 \times 512$ images for facilitating the research on makeup transfer.
It is demonstrated that by using high-resolution images, our PSGAN++ can produce high-quality results.
Fourth, we add considerable new experimental results including ablation study, model setting, and visualization analysis to fully validate our methodology.
\vspace{-4mm}
\section{Related Work}
\subsection{Makeup Transfer and Removal}
Makeup transfer has been studied a lot over the past years \cite{Tong2007ExampleBasedCT,Guo2009DigitalFM,Li2015SimulatingMT,Liu2016MakeupLA,Liu2014WowYA,Alashkar2017ExamplesRulesGD}.
BeautyGAN~\cite{Li2018BeautyGANIF} first applies a GAN framework with dual inputs and outputs for makeup transfer.
It uses a makeup region loss that matches the color histogram in different parts of the face for instance-level makeup transfer.
BeautyGlow \cite{ChenBeautyGlow2019} adopts a similar idea of the Glow framework and decomposes makeup and non-makeup components.
PairedCycleGAN~\cite{Chang2018PairedCycleGANAS} employs an additional discriminator to guide makeup transfer using pseudo transferred images generated by warping the reference face to the source face.
LADN~\cite{Gu2019LADNLA} leverages additional multiple overlapping local discriminators for dramatic makeup transfer.

Makeup removal is a blind reverse of the unknown beautification process and restores faces similar to what are captured by cameras.
Makeup-Go~\cite{chen2017makeup} analyzes the component domination effect in blind reversion and ameliorates it with a component regression network.
PairedCycleGAN~\cite{Chang2018PairedCycleGANAS} leverages an asymmetric framework that contains an additional removal sub-network for makeup removal.
DRL~\cite{li2019disentangled} disentangles the identity information and makeup style for makeup transfer and removal.

\blue{
\textbf{Superiority of PSGAN++ over existing methods. }
The aforementioned makeup transfer approaches are effective, but PSGAN++ performs better in the following aspects.
First, only PSGAN and PSGAN++ support pose/expression robust transfer because of the novel AMM module.
Second, traditional methods do not provide the partial transfer function. Thanks to the MDNet module which encodes makeup information into spatial-aware makeup matrices, PSGAN series methods enable  partial transfer.
Third, degree-controllability is gradually becoming a common feature of recent makeup transfer methods including BGlow\cite{ChenBeautyGlow2019}, LADN\cite{Gu2019LADNLA}, PSGAN, and PSGAN++. 
Fourth,  PSGAN++ is able to transfer makeup details, e.g., highlight and blush, which distinguishes PSGAN++ from other existing makeup transfer methods.
Last but not least, PGAN\cite{Chang2018PairedCycleGANAS}, LADN\cite{Gu2019LADNLA}  and PSGAN++ can remove makeup. 
}

\begin{figure*}[!t]
    \centering
    \includegraphics[width=0.9\linewidth]{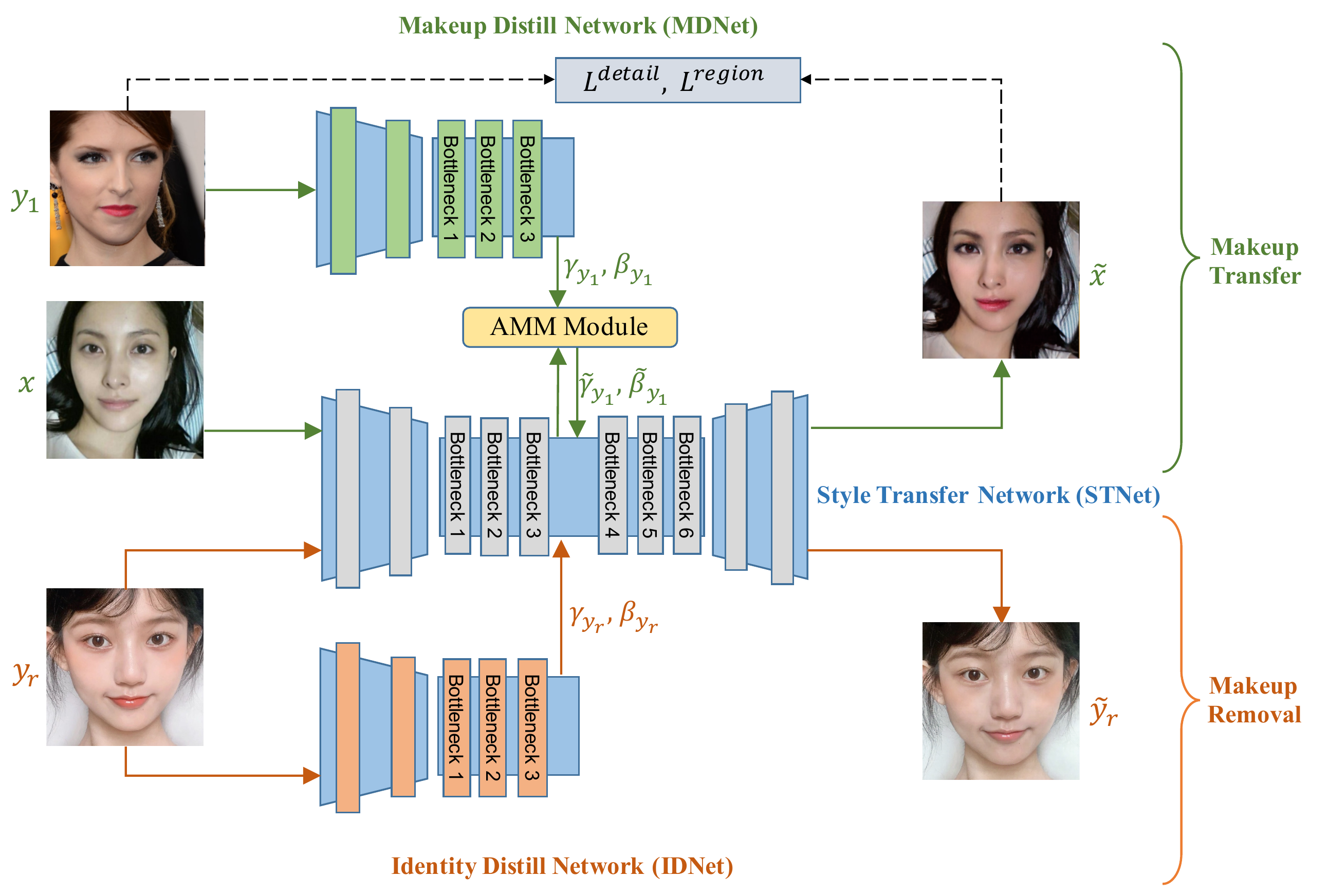}
    \caption{
        Illustration of the workflow of our PSGAN++.
        The PSGAN++ can transfer and remove makeup.
        For makeup transfer, PSGAN++ takes the source image $x$ and the reference image $y_1$ as input to produce transferred image $\tilde{x}$. The MDNet distills makeup matrices $\gamma_{y_1}$ and $\beta_{y_1}$ from the reference image, and the AMM module applies the adapted makeup matrices $\tilde{\gamma}_{y_1}$ and $\tilde{\beta}_{y_1}$ to the decoder part of STNet to achieve makeup transfer.
        For makeup removal, PSGAN++ takes the with-makeup image $y_r$ as input to produce de-makeup image $\tilde{y}_r$. The IDNet extracts the identify matrices $\gamma_{y_r}$ and $\beta_{y_r}$ from $y_r$, which are then fed to the STNet for makeup removal.
        }
    \label{framework}
    \vspace{-5mm}
\end{figure*}

\vspace{-3mm}
\subsection{Style Transfer}
Style transfer has been investigated extensively \cite{Gatys2015ANA,Gatys2016PreservingCI,Johnson2016PerceptualLF,Luan2017DeepPS,Taigman2016UnsupervisedCI,DRIT,DRIT_plus}.
Gatys et al. \cite{Gatys2016ImageST} proposed to derive image representations from CNN, which can be separated and recombined to synthesize images.
Some methods tackle the fast style transfer problem.
DRIT and DRIT++ \cite{DRIT,DRIT_plus} introduce a disentangled representation framework for image-to-image translation by applying a content discriminator to facilitate the factorization of the domain-invariant content space and domain-specific attribute space.
Dumoulin et al.~\cite{Dumoulin2016ALR} found the vital role of normalization in style transfer networks and achieved fast style transfer by the conditional instance normalization.
However, their method can only transfer a fixed set of styles and cannot adapt to arbitrary new styles.
Then, Huang et al.~\cite{Huang2017ArbitraryST} proposed adaptive instance normalization (AdaIN) that aligns the mean and variance of the content features with those of the style features for arbitrary style transfer.
Here, our focus is placed upon spatial-aware makeup transfer for each pixel rather than transferring a general style from the reference.

\vspace{-4mm}
\subsection{Attention Mechanism}
Attention mechanism has been widely applied to image generation \cite{yu2018generative,xu2018attngan,gao2020adversarialnas,zhu2019progressive}.
Vaswani et al.~\cite{Vaswani2017AttentionIA} first proposed the attention mechanism in the natural language processing area, which computes the response at a position in a sequence by attending to all positions and taking their weighted average in the embedding space.
AttnGAN~\cite{xu2018attngan} applies an attention mechanism to the text-to-image generation task to synthesize fine-grained details at each sub-region by paying attention to relevant words.
Yu et al.~\cite{yu2018generative} designed a contextual attention method for image inpainting to employ surrounding image patches as informative clues for a better generation.
PATN~\cite{zhu2019progressive} uses a progressive pose attention module for pose-guided person generation and obtains obvious improvements.
Wang et al.~\cite{Wang2017NonlocalNN} proposed a non-local network that computes the response at a position by taking a weighted sum of the features at all positions.
Unlike the non-local network that only considers visual appearance similarities, our proposed AMM module computes the weighted sum of another feature map by considering both visual appearances and locations.

\vspace{-3mm}
\section{PSGAN++}
\subsection{Task Definition}
Let $X$ and $Y$ denote the set of source images and that of reference images, respectively.
Note that paired datasets are not required. That is, the source and reference images may belong to different identities.
$x$ is sampled from $X$ according to the distribution $\mathcal{P}_{X}$, and $y_1$, $y_r$ are sampled from $Y$ according to the distribution $\mathcal{P}_{Y}$.
The PSGAN++ learns a makeup transfer function $\tilde{x} = T(x,y_1)$, where the transferred image $\tilde{x}$ has the makeup style of the reference image $y_1$ and preserves the identity of the source image $x$.
In addition, it learns a makeup removal function $\tilde{y}_r = R(y_r) $, which removes the makeup from $y_r$ to produce the de-makeup image $\tilde{y}_r$.

\vspace{-3mm}
\subsection{Framework}
\textbf{Overall.} The overall framework of PSGAN++ is shown in Figure \ref{framework}.
PSGAN++ is designed to address two tasks, makeup transfer and makeup removal.
For makeup transfer, PSGAN++ takes the reference image $y_1$ and the source image $x$ as inputs.
The Makeup Distill Network (MDNet) extracts and represents the makeup style from  $y_1$ as two makeup matrices $\gamma_{y_1}$ and $\beta_{y_1}$.
Since the source and reference images may have large discrepancies in poses and expressions, the extracted makeup matrices cannot be directly applied to the source image $x$.
We propose an Attentive Makeup Morphing (AMM) module to morph them to two new matrices $\tilde{\gamma}_{y_1}$ and $\tilde{\beta}_{y_1}$ which are adaptive to the source image by considering the pixel-wise similarities of the source and reference.
Then, the adaptive makeup matrices $\tilde{\gamma}_{y_1}$ and $\tilde{\beta}_{y_1}$ are applied to the decoder part of the Style Transfer Network (STNet) to perform makeup transfer by element-wise multiplication and addition.
With the spatial-aware makeup matrices and the AMM module, PSGAN++ can achieve partial, degree-controllable, and robust makeup transfer.

Our PSGAN++ can also remove the makeup automatically.
To remove the makeup from the with-makeup input image $y_r$, the Identity Distill Network (IDNet) disentangles and represents the identity information from it as identity matrices $\gamma_{y_r}$ and $\beta_{y_r}$.
Unlike makeup transfer, there is no misalignment problem in makeup removal, thus the AMM module is not needed here.
The obtained identity matrices and $y_r$ are fed into the STNet to produce the de-makeup image $\tilde{y}_r$.

\begin{figure}[!t]
    \centering
    \includegraphics[width=1\linewidth]{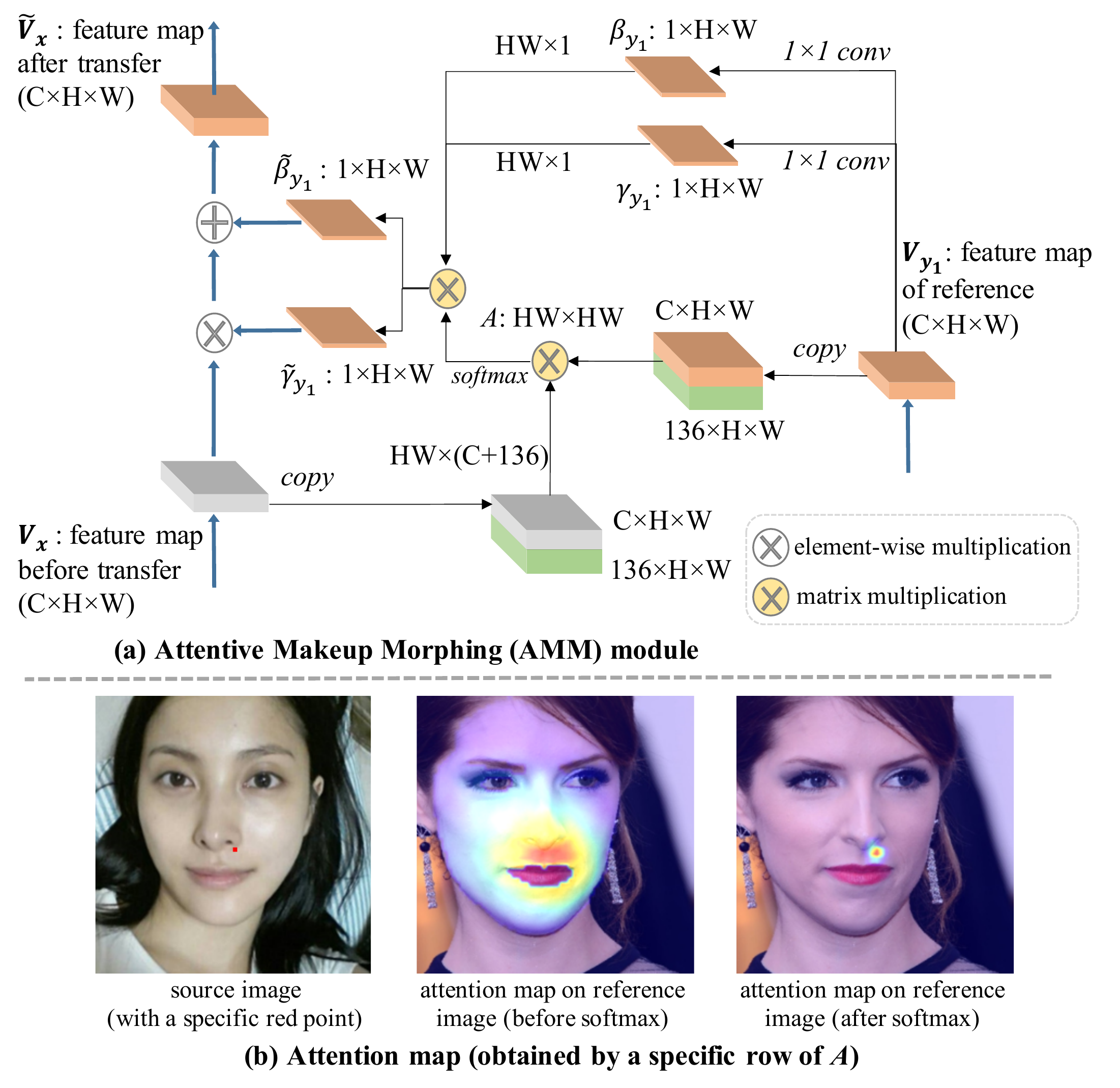}
    \caption{
        \textbf{(a)} Illustration of AMM module. Green blocks with $136$ ($68 \times 2$) channels indicate relative position features of the pixels, which are then concatenated with $C$-channel visual features. Thus, the attention map is computed for each pixel in the source image based on the similarities of relative positions and visual appearances. The adapted makeup matrices $\tilde{\gamma}_{y_1}$ and $\tilde{\beta}_{y_1}$ are produced by the AMM module, which are then multiplied and added to feature maps of MANet element-wisely. The orange and gray blocks indicate visual features with makeup and without makeup respectively. \textbf{(b)} Attention maps for a specific red point in the source image. Note that we only calculate attentive values for pixels that belong to the same facial region. Thus, there are no response values on the lip and eye of the reference image.
    }
    \vspace{-9mm}
    \label{AMM}
\end{figure}

\textbf{Makeup distill network (MDNet).} The MDNet utilizes an encoder-bottleneck architecture.
It disentangles the makeup-related features (e.g. lip gloss, eye shadow) from the intrinsic facial features (e.g. facial shape, size of eyes).
The makeup related features are represented as two makeup matrices $\gamma_{y_1}$ and $\beta_{y_1}$, which are used to transfer the makeup by pixel-level operations. As shown in Figure \ref{AMM}(a), the output feature map of MDNet $\mathbf{V_{y_1}}\in \mathbb{R}^{C \times H \times W}$ is fed into two $1\times1$ convolution layers to produce $\gamma_{y_1} \in \mathbb{R}^{1\times H \times W}$ and $\beta_{y_1} \in \mathbb{R}^{1 \times H \times W}$, where $C$, $H$ and $W$ are the number of channels, height and width of the feature map.

\textbf{Attentive makeup morphing (AMM) module.}
Since the source and reference images may have different poses and expressions, the obtained spatial-aware $\gamma_{y_1}$ and $\beta_{y_1}$ cannot be applied directly to the feature map of the source image.
The AMM module is to calculate an attentive matrix $A \in \mathbb{R}^{HW \times HW}$ to specify how a pixel in the source image $x$ is morphed from the pixels in the reference image $y_1$, where $A_{i,j}$ indicates the attentive value between the $i$-th pixel $x^i$ in the feature map of $x$ and the $j$-th pixel $y^j_1$ in the feature map of $y_1$.
Intuitively, makeup should be transferred between the pixels with similar relative positions on the face, and the attentive values between these pixels should be high.
For example, the lip gloss region of the transferred result $\tilde{x}$ should be sampled from the corresponding lip gloss region of the reference image $y_1$.

To describe the relative positions, we take the $68$ facial landmarks as anchor points.
The relative position features of $x^i$ and $y_1^j$ are represented by $\mathbf{p}^i_x \in \mathbb{R}^{136}$ and $\mathbf{p}^j_{y_1} \in \mathbb{R}^{136}$, which are reflected in the differences of coordinates between pixels and the $68$ facial landmarks.
Specifically, $\mathbf{p}^i_x$ and $\mathbf{p}^j_{y_1}$ are defined as
\begin{equation}
    \small
    \begin{aligned}
    \mathbf{p}^i_x = [&f(x^i)-f(l^1_x), f(x^i)-f(l^2_x), \dots, f(x^i) - f(l^{68}_x),\\ &g(x^i)-g(l^1_x), g(x^i)-g(l^2_x), \dots, g(x^i) - g(l^{68}_x)], \\
    \mathbf{p}^j_{y_1} = [&f(y_1^j)-f(l^1_{y_1}), f(y_1^j)-f(l^2_{y_1}), \dots, f(y_1^j) - f(l^{68}_{y_1}),\\ &g(y_1^j)-g(l^1_{y_1}), g(y_1^j)-g(l^2_{y_1}), \dots, g(y_1^j) - g(l^{68}_{y_1})],
    \end{aligned}
    \label{p}
\end{equation}
where $f(\cdot)$ and $g(\cdot)$ indicate the coordinates on horizontal and vertical axes, $l^k_x$ and $l^k_{y_1}$ indicate the $k$-th facial landmark of $x$ and $y_1$ obtained by the 2D facial landmark detector \cite{Zhang2016JointFD}.
The facial landmarks serve as the anchor points when calculating $\mathbf{p}$.
Besides, in order to handle faces of different sizes in the images, we divide $\mathbf{p}$ by its two-norm (i.e., $\frac{\mathbf{p}}{\left\| \mathbf{p} \right\|}$) when calculating the attentive matrix.

Moreover, to avoid unreasonable sampling pixels with similar relative positions but different semantics, we also consider the visual similarities between pixels.
The visual similarities between $x^i$ and $y^j_1$ are denoted as the similarities between their corresponding visual features $\mathbf{V}_x^i$ and $\mathbf{V}^j_{y_1}$ that are extracted from the third bottleneck of MANet and MDNet respectively.
To lift the priority of the relative position, we multiply the visual features by a relatively small weight $w$ when calculating $A$.
Then, the relative position features are resized and concatenated with the visual features along the channel dimension.
As Figure \ref{AMM} (a) shows, the attentive value $A_{i,j}$ is computed by considering the similarities of both visual appearances and relative positions via
\begin{equation}
\small
A_{i, j} = \frac{\exp \left( [w\mathbf{V}^i_x, \frac{\mathbf{p}^i_x}{\left\| \mathbf{p}^i_x \right\|} ]^T [w\mathbf{V}^j_{y_1}, \frac{\mathbf{p}^j_{y_1}}{\left\| \mathbf{p}^j_{y_1} \right\|}]  \right) \mathbb{I}  (m^i_x = m^j_{y_1}) }{\sum_j \exp \left( [w\mathbf{V}^i_x, \frac{\mathbf{p}^i_x}{\left\| \mathbf{p}^i_x \right\|} ]^T [w\mathbf{V}^j_{y_1}, \frac{\mathbf{p}^j_{y_1}}{\left\| \mathbf{p}^j_{y_1} \right\|}] \right) \mathbb{I}  (m^i_x = m^j_{y_1})}.
\label{equ5}
\end{equation}
Here $[\cdot, \cdot]$ denotes the concatenation operation, $\mathbf{V}_x^i, \mathbf{V}^j_{y_1} \in \mathbb{R}^{C}$ and $\mathbf{p}^i_x, \mathbf{p}^j_{y_1} \in \mathbb{R}^{136} $ indicate the visual features and relative position features, and $w$ is the weight for visual features.
$\mathbb{I}(\cdot)$ is an indicator function whose value is $1$ if the inside formula is true; $m_x, m_{y_1} \in \{0, 1, \dots, N-1\}^{H \times W}$ are the face parsing maps of the source image $x$ and the reference image $y_1$, where $N$ stands for the number of facial regions ($N$ is $3$ in our experiments including eyes, lip and skin); $m^i_x$ and $m^j_{y_1}$ indicate the facial regions that $x^i$ and $y^j_1$ belong to.
Note that we only consider the pixels belonging to the same facial region, i.e., $m^i_x = m^j_{y_1}$ , by applying the indicator function $\mathbb{I}(\cdot)$.

Given a specific point marked in red in the lower-left corner of the nose in the source image,
the middle image of Figure \ref{AMM} (b) shows its attention map by reshaping a specific row of the attentive matrix $A_{i,:} \in \mathbb{R}^{1 \times HW}$ to $H \times W$.
We can see that only the pixels around the left corner of the nose have large values.
After applying softmax, attentive values become more gathered.
It verifies that our proposed AMM module can locate and attend to semantically similar pixels.
We multiply the attentive matrix $A$ by the $\gamma_{y_1}$ and $\beta_{y_1}$ to obtain the morphed makeup matrices $\tilde{\gamma}_{y_1}$ and $\tilde{\beta}_{y_1}$.
More specifically, they are computed by
\begin{equation}
\small
    \begin{split}
    {\tilde{\gamma}^i_{y_1}} = \sum_{j} A_{i, j} \gamma^j_{y_1}, \\
    {\tilde{\beta}^i_{y_1}} = \sum_{j} A_{i, j} \beta^j_{y_1}, \\
    \label{equ4}
    \end{split}
\end{equation}
where $i$ and $j$ are the pixel index of $x$ and $y_1$.
After that, the matrix $\tilde{\gamma}_{y_1}  \in \mathbb{R}^{1 \times H \times W}$ and $\tilde{\beta}_{y_1} \in \mathbb{R}^{1 \times H \times W}$ are duplicated and expanded along the channel dimension to produce the makeup tensors $\mathbf{\tilde{\Gamma}_{y_1}}  \in \mathbb{R}^{C \times H \times W}$ and $\mathbf{\tilde{B}_{y_1}} \in \mathbb{R}^{C \times H \times W}$, which will be input to STNet for makeup transfer.

\textbf{Identity distill network (IDNet).}
The IDNet utilizes the same encoder-bottleneck architecture as the MDNet, to distill the intrinsic facial features.
The identity related features of $y_r$ in Figure \ref{framework} are represented as two identity matrices $\gamma_{y_r}$ and $\beta_{y_r}$.
After that, the matrices $\gamma_{y_r}  \in \mathbb{R}^{1 \times H \times W}$ and $\beta_{y_r} \in \mathbb{R}^{1 \times H \times W}$ are duplicated and expanded along the channel dimension to produce the identity tensors $\mathbf{\Gamma_{y_r}}  \in \mathbb{R}^{C \times H \times W}$ and $\mathbf{B_{y_r}}  \in \mathbb{R}^{C \times H \times W}$, which will be input to STNet for makeup removal.

\textbf{Style transfer network (STNet).} The STNet utilizes an encoder-bottleneck-decoder architecture.
As shown in Figure \ref{framework}, the encoder part of STNet shares the same architecture with MDNet and IDNet, but they do not share parameters.
In the encoder part, we use instance normalization without affine parameters that make the feature map to be a normal distribution.
For \emph{makeup transfer}, the makeup tensors are applied to the source image feature map in the bottleneck part.
the activation values of the transferred feature map $\mathbf{\tilde{V}_{x}}$ are calculated by
\begin{equation}
\mathbf{\tilde{V}_{x}} = \mathbf{\tilde{\Gamma}_{y_1}} \mathbf{V_{x}} + \mathbf{\tilde{B}_{y_1}},
\label{equ1}
\end{equation}
where $\mathbf{V_{x}}\in \mathbb{R}^{C \times H \times W}$ is the visual feature map of $x$. $\mathbf{\tilde{\Gamma}_{y_1}}$ and $\mathbf{\tilde{B}_{y_1}}$ are makeup tensors extracted from $y_1$.
The updated feature map $\mathbf{\tilde{V}_{x}}$ is then fed to the subsequent decoder part of STNet to produce the transferred result.
Similarly, the \emph{makeup removal} can be achieved by
\begin{equation}
\mathbf{\tilde{V}_{y_r}} = \mathbf{\Gamma_{y_r}} \mathbf{V_{y_r}} + \mathbf{B_{y_r}},
\label{equ2}
\end{equation}
where $\mathbf{\Gamma_{y_r}}$ and $\mathbf{B_{y_r}}$ are identity tensors extracted from $y_r$ using IDNet, and $\mathbf{V_{y_r}}\in \mathbb{R}^{C \times H \times W}$ is the visual feature map of $y_r$.
The updated feature map $\mathbf{\tilde{V}_{y_r}}$ is also fed to the subsequent decoder part of STNet to produce the de-makeup image $\tilde{y}_r$.

\subsection{Objective Function}
\textbf{Adversarial loss.} We utilize two discriminators $D_X$ and $D_Y$ for the source image domain $X$ and the reference image domain $Y$, discriminating between generated images and real images and thus helping generators synthesize realistic outputs.

\blue{
$D_X$ and $D_Y$ are two different discriminators with the same PatchGANs architecture \cite{isola2017image} to classify the $70 \times 70$ overlapping image patches. 
$D_X$ discriminates whether the images belong to the source image domain $X$ (non-makeup image domain).
In contrast, $D_Y$ discriminates whether the images belong to the reference image domain $Y$ (with-makeup image domain).
}

The adversarial losses $L_D^{adv}$, $L_G^{adv}$ for discriminator and generator are computed by
\begin{equation}
    \begin{gathered}
        \begin{aligned}
            L_{D}^{adv} &= - \mathbb{E}_{x \sim \mathcal{P}_{X}}\left[\log D_{X}(x)\right] - \mathbb{E}_{y_1 \sim \mathcal{P}_{Y}}\left[\log D_{Y}(y_1)\right] \\ &- \mathbb{E}_{y_r \sim \mathcal{P}_{Y}}\left[\log \left(1-D_{X}(R(y_r))\right)\right] \\ &- \mathbb{E}_{x \sim \mathcal{P}_{X}, y_1 \sim \mathcal{P}_{Y}}\left[\log \left(1-D_{Y}(T(x, y_1))\right)\right], \\
            L_{G}^{adv} &= - \mathbb{E}_{y_r \sim \mathcal{P}_{Y}}\left[\log \left(D_{X}(R(y_r))\right)\right] \\ &- \mathbb{E}_{x \sim \mathcal{P}_{X}, y_1 \sim \mathcal{P}_{Y}}\left[\log \left(D_{Y}(T(x, y_1 ))\right)\right].
        \end{aligned}
    \end{gathered}
\end{equation}

\begin{figure}[!t]
    \centering
    \includegraphics[width=1.0\linewidth]{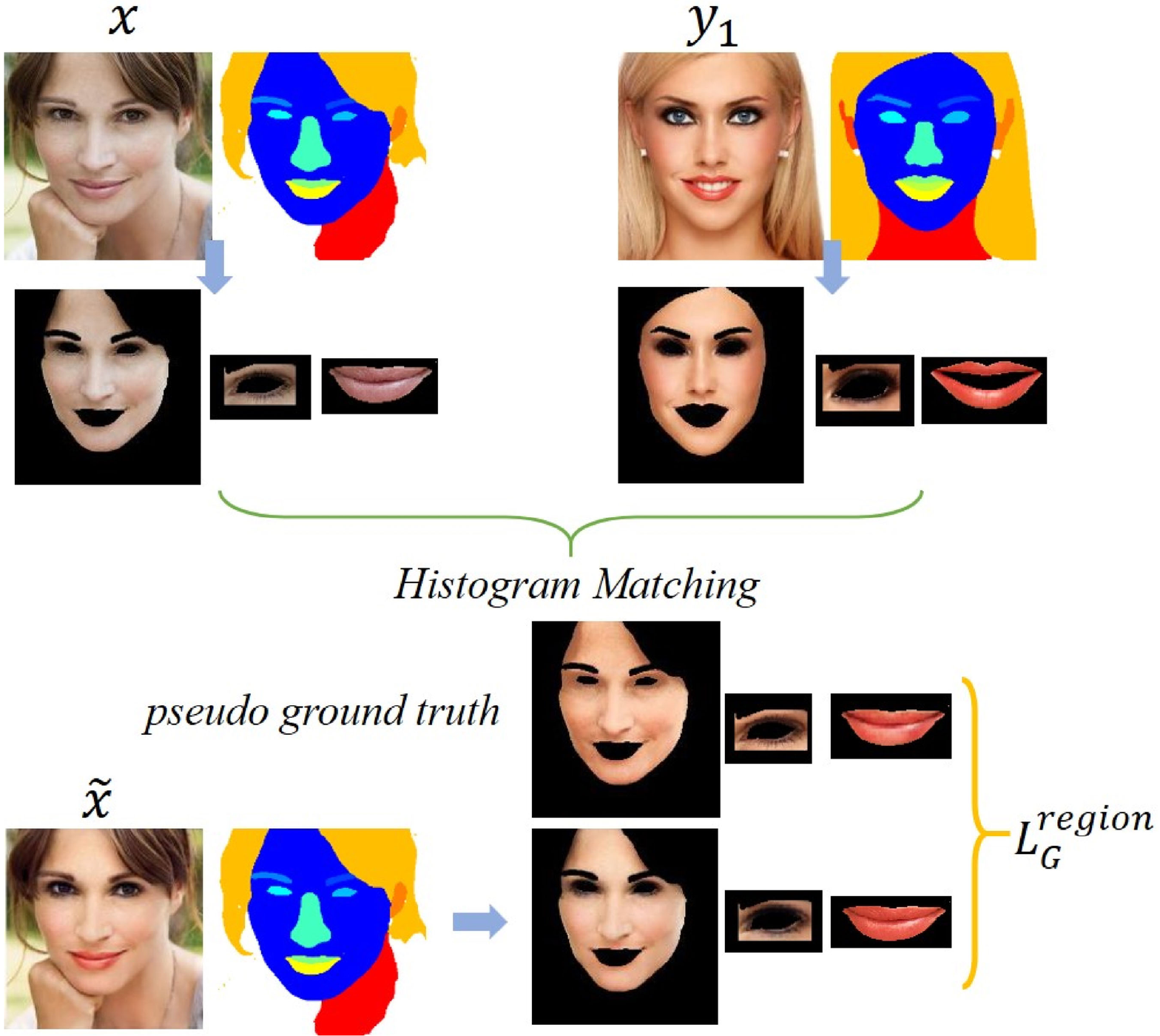}
    \blue{
    \caption{
        Details of the makeup region loss.
    }
    \vspace{-5mm}
    \label{hm}
    }
\end{figure}

\textbf{Cycle consistency loss.} Due to the lack of triplet data (source image, reference image, and transferred image), we train the network in an unsupervised way.
Here, we introduce the cycle consistency loss proposed by \cite{Zhu2017UnpairedIT}.
We use the L1 loss to constrain the reconstructed images and define the cycle consistency loss $L_G^{cyc}$ as
\begin{equation}
    \begin{aligned}
        L_G^{cyc} &= \mathbb{E}_{x \sim \mathcal{P}_{X}, y_1 \sim \mathcal{P}_{Y}}\left[ \left\| R(T(x, y_1)) - x \right\|_{1} \right]  \\ &+\mathbb{E}_{y_r \sim \mathcal{P}_{Y}}\left[ \left\| T(R(y_r), y_r) - y_r \right\|_{1} \right].
    \end{aligned}
\end{equation}

\textbf{Perceptual loss.} When transferring the makeup style, the transferred image is required to preserve the personal identity.
Instead of directly measuring differences at pixel-level, we utilize a VGG-16 model pre-trained on ImageNet to compare the activations of source images and generated images in the hidden layer.
Let $F_l(\cdot)$ denote the output of the $l$-th layer of the VGG-16 model.
We introduce the perceptual loss $L_G^{per}$ to measure their differences using L2 loss:
\begin{equation}
    \begin{aligned}
        L_G^{per} &= \mathbb{E}_{x \sim \mathcal{P}_{X}, y_1 \sim \mathcal{P}_{Y}}\left[ \left\| F_l(T(x, y_1)) - F_l(x) \right\|_{2} \right]  \\ &+\mathbb{E}_{y_r \sim \mathcal{P}_{Y}}\left[ \left\| F_l(R(y_r)) - F_l(y_r) \right\|_{2} \right].
    \end{aligned}
\end{equation}

\textbf{Makeup region loss.} To provide coarse guidance for makeup transfer, we utilize the makeup region loss proposed by \cite{Li2018BeautyGANIF}. 
\blue{
    The details of the makeup region loss are shown in Figure \ref{hm}. The inputs include the source image $x$, reference image $y_1$,  transferred image $\tilde{x}$ and their parsing masks including three facial regions (lip gloss, skin, and eye shadow).
    We first apply histogram matching on the three facial regions separately to obtain the pseudo ground truth of each facial region.
    Then we recombine them into $HM(x, y_1)$. 
    Finally, we use $HM(x, y_1)$ as pseudo ground truth and calculate the makeup region loss $L^{region}_G$ between $HM(x, y_1)$ and transferred image $\tilde{x}$.
}

\blue{
    Histogram matching manipulates the pixels of the source image $x$ so that its histogram matches the histogram of the reference image $y_1$.
    For RGB images, the matching is done independently for each channel.
    The output image has a similar hue (i.e., histograms) with $y_1$ but maintains the content of $x$, thus can be used as pseudo ground truth.
    Specifically, given $x$ and $y_1$, we compute their histograms and calculate the cumulative distribution functions of the two images' histograms as $F_{x}$ and $F_{y_1}$. Then for each gray level $G_{x} \in [0, 255]$, we find the gray level $G_{y_1}$, for which $F_{x}(G_{x})=F_{y_1}(G_{y_1})$. Thus we get the histogram matching function: $M(G_{x})=G_{y_1}$. Finally, we apply the function $M(\cdot)$ on each pixel of $x$ to obtain the output pseudo ground truth $HM(x, y_1)$.
}

\blue{
    Overall, we use $HM(x, y_1)$ as a kind of pseudo ground truth and calculate the makeup region loss $L_G^{region}$ as coarse guidance by:
}
\begin{equation}
    \begin{aligned}
        L_G^{region} &= \mathbb{E}_{x \sim \mathcal{P}_{X}, y_1 \sim \mathcal{P}_{Y}}\left[ \left\| T(x, y_1) - HM(x, y_1) \right\|_{2} \right].
    \end{aligned}
\end{equation}

\begin{figure}[!t]
    \centering
    \includegraphics[width=0.8\linewidth]{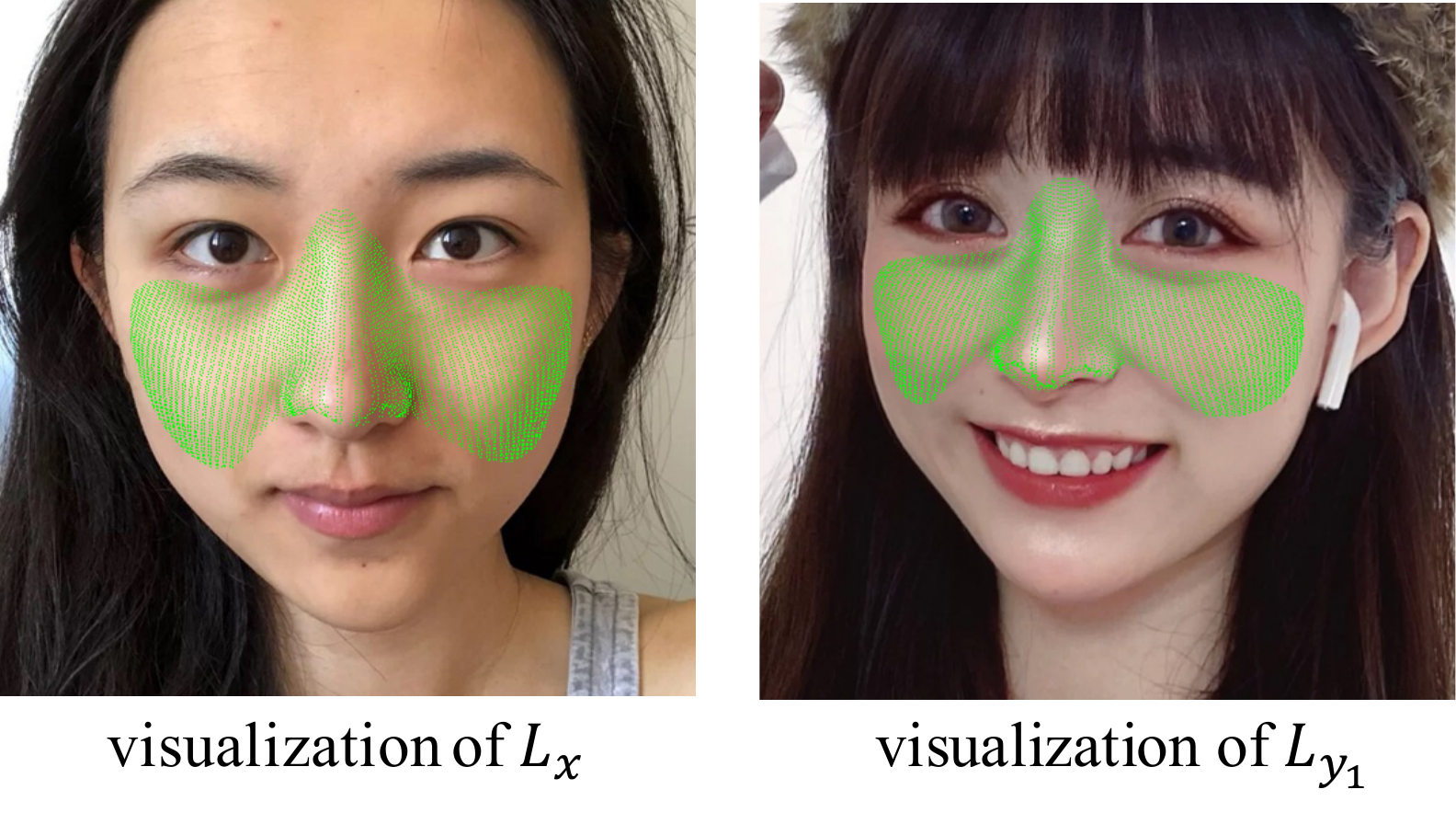}
    \caption{
        Visualization of the selected $K$ landmarks around nose and cheeks.
    }
    \vspace{-3mm}
    \label{detailed_vis}
\end{figure}

\textbf{Makeup detail loss.}
The makeup region loss provides supervision at the facial region level, but such a loss can hardly transfer makeup details including highlight and blush.
To address these problems, our PSGAN++ applies a novel makeup detail loss.
To compute the makeup detail loss $L_G^{detail}$, we first leverage a dense face alignment method \cite{feng2018prn} to detect dense facial landmarks for the source image $x$ and the reference image $y_1$.
We then select $K$ landmarks that lie in the area of makeup details, i.e., nose and cheeks, to form the makeup detail landmarks $L_x$ and $L_{y_1}$.
The visualization of the selected landmarks is shown in Figure \ref{detailed_vis}, the number of $K$ is $20560$ in our experiments.
The makeup detail loss is to calculate the difference between the transferred image $T(x, y_1)$ and the reference image $y_1$ in the corresponding makeup detail landmarks via
\begin{equation}
    \begin{gathered}
    L_G^{detail} = \mathbb{E}_{x \sim \mathcal{P}_{X}, y_1 \sim \mathcal{P}_{Y}} \left[ \sum_{k} \|T(x, y_1)^k - y^k_1 \|_{1} \right],
    \end{gathered}
\end{equation}
where $T(x, y_1)^k$ and $y^k_1$ indicate the pixel values of the $k$-th landmarks in $T(x, y_1)$ and $y_1$.
By directly computing the differences between corresponding landmarks as supervision, PSGAN++ learns to generate the same makeup details as the reference.

\textbf{Total loss.} The loss $L_D$ and $L_G$ for discriminator and generator of our approach can be expressed as
\begin{equation}
        \begin{aligned}
            L_D &= \lambda_{adv} L_D^{adv}, \\
            L_G &= \lambda_{adv} L_G^{adv} + \lambda_{cyc} L_G^{cyc} + \lambda_{per} L_G^{per} \\
            &+ \lambda_{region} L_G^{region} + \lambda_{detail} L_G^{detail},
        \end{aligned}
\end{equation}
where $\lambda_{adv}$, $\lambda_{cyc}$, $\lambda_{per}$, $\lambda_{region}$, $\lambda_{detail}$ are the weights to balance the multiple objectives.

\section{Experiments}

\begin{figure*}[t]
    \includegraphics[width=1\linewidth]{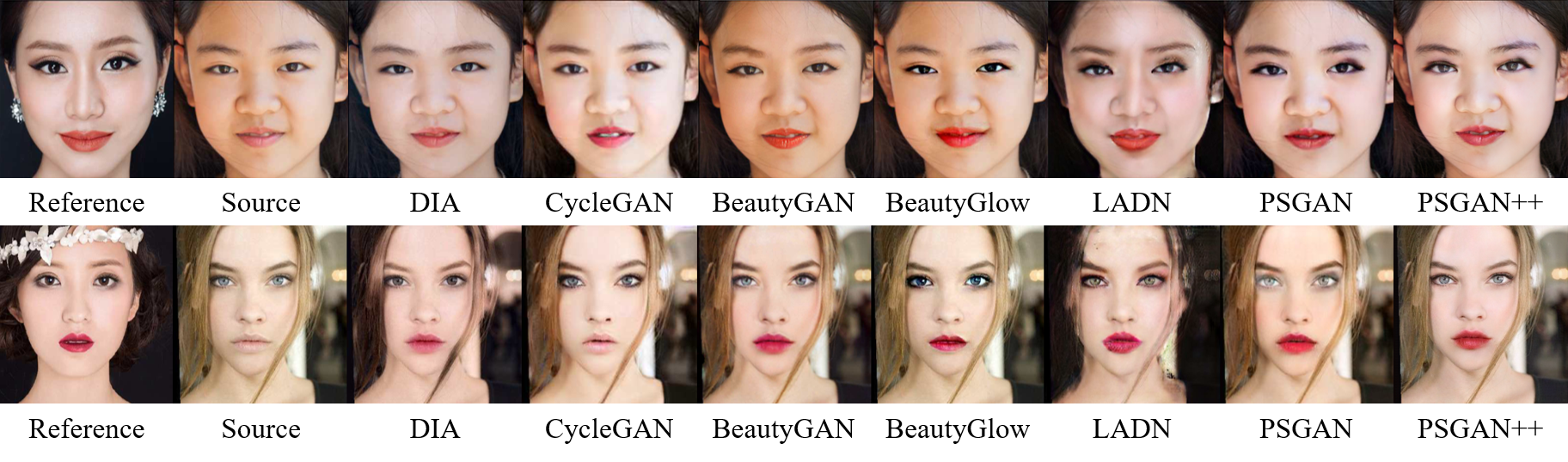}
    \caption{Qualitative comparison over frontal faces and neutral expressions. PSGAN++ is able to preserve the face structure when transferring the makeup.}
    \label{res1}
    \vspace{-3mm}
\end{figure*}

\subsection{Experimental Setting}

\textbf{Datasets.}  One public dataset and two newly collected datasets are used in our experiments.
\emph{Makeup Transfer} (MT) dataset \cite{Li2018BeautyGANIF} is a widely used makeup dataset containing $1,115$ non-makeup images and $2,719$ with-makeup images with resolution of $361 \times 361$.
\emph{Makeup Transfer in the Wild} (MT-Wild) is a newly collected dataset that contains images with various poses and expressions with complex backgrounds.
In concrete, we collect images from the Internet and discard those with frontal faces and neutral expressions manually.
Finally, $403$ with-makeup images and $369$ non-makeup images are remained and cropped and resized to $256 \times 256$.
\emph{Makeup Transfer High Resolution} (MT-HR) dataset is the other newly collected high-resolution makeup dataset with $2,000$ with-makeup images and $1,000$ non-makeup images in $512 \times 512$, which is used to evaluate how various methods can handle high-resolution images.
We randomly select $200$ with-makeup images and $100$ non-makeup images for testing and the remaining images are used for training and validation.

\begin{table}[!t]
    \centering
    \small
    \setlength{\tabcolsep}{1mm}{
        \begin{tabular}{@{}ccccccc@{}}
            \toprule
            Test set & BGAN & DIA & CGAN & LADN  & PSGAN & PSGAN++\\ \midrule
            MT     & 12.50 & 4.00 & 2.00  &  0.50 & 33.50 & \textbf{47.50} \\ \midrule
            MT-Wild     & 7.25 &  3.25 & 1.50  &  0.25  & 37.50  & \textbf{50.25}\\ \midrule
            MT-HR  & 5.75 & 2.75 & 1.50 & 0.25 & 38.25 & \textbf{51.50} \\ \bottomrule
    \end{tabular}}
    \caption{Ratio selected as best (\%).}
    \label{t2}
    \vspace{-5mm}
\end{table}

\textbf{Implementation details.}
For a low-resolution setting ($256 \times 256$), we train and test our network on the MT dataset, following the splitting strategy of \cite{Li2018BeautyGANIF}.
In order to prove our PSGAN++ has the ability to tackle pose and expression differences, we further test our network on the MT-Wild dataset.
Note that MT-Wild is only used for testing, and the model is still trained on the training part of the MT dataset.
For the high-resolution setting ($512 \times 512$), we train and test our network on the MT-HR dataset.
We resize the images to $256 \times 256$ for the low-resolution setting (MT and MT-Wild) and $512 \times 512$ for the high-resolution setting (MT-HR).
When calculating the perceptual loss, we use the $relu\_4\_1$ feature layer of VGG-16.
We optimize the PSGAN++ with $\lambda_{adv} = 1$, $\lambda_{cyc} = 10$, $\lambda_{per} = 0.005$, $\lambda_{region} = 1$, $\lambda_{detail} = 3$, and the weight for visual features in AMM is set to $0.01$.
We train the network for 50 epochs using Adam optimizer \cite{kingma2014adam} with the learning rate of $0.0002$ and the batch size of 1.

\subsection{Comparison with SOTA}

We compare with the baselines including general image-to-image translation methods DIA~\cite{Liao2017VisualAT} and CycleGAN~\cite{Zhu2017UnpairedIT} as well as state-of-the-art makeup transfer methods BeautyGAN (BGAN)~\cite{Li2018BeautyGANIF}, PairedCycleGAN (PGAN)~\cite{Chang2018PairedCycleGANAS}, BeautyGlow (BGlow)~\cite{ChenBeautyGlow2019} and LADN~\cite{Gu2019LADNLA}.
Current makeup transfer methods leverage either face parsing maps \cite{Chang2018PairedCycleGANAS,ChenBeautyGlow2019,Li2018BeautyGANIF} or facial landmarks \cite{Gu2019LADNLA} for training.

\begin{figure}[t]
    \includegraphics[width=1\linewidth]{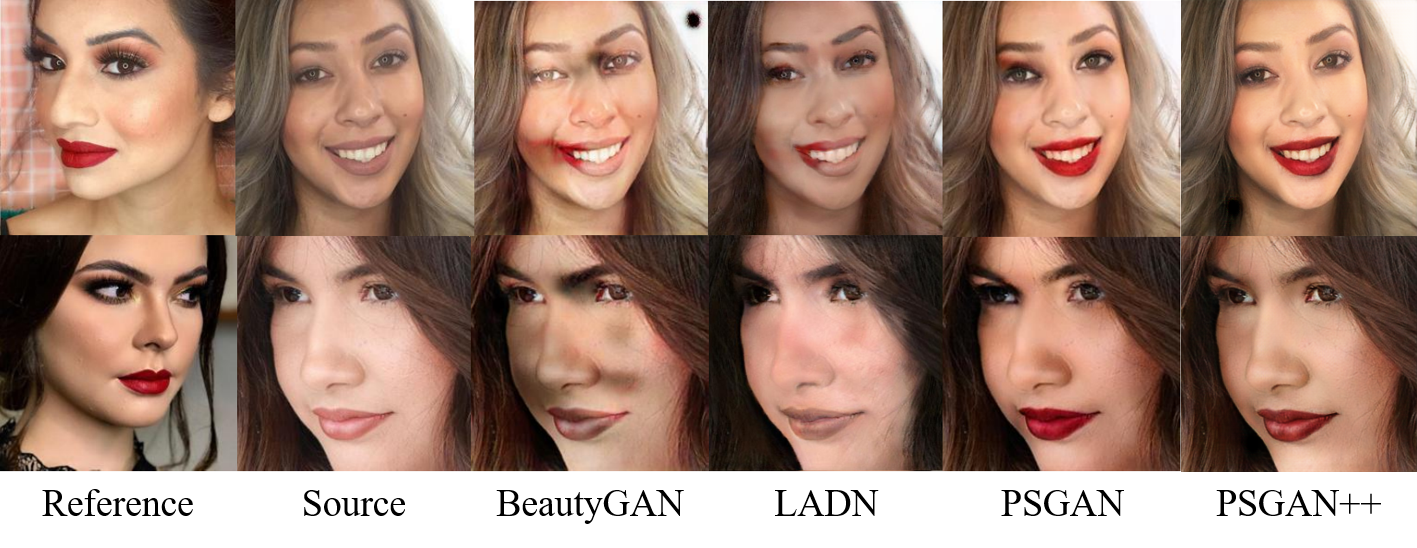}
    \caption{Qualitative comparison over different poses and expressions. PSGAN++ is able to realize makeup transfer regardless of different poses.}
    \label{wild}
    \vspace{-3mm}
\end{figure}

\textbf{User studies} are conducted for quantitative comparisons among six makeup transfer methods, including BGAN, CGAN, DIA, LADN, PSGAN, and PSGAN++ on Amazon Mechanical Turk platform\footnote{https://www.mturk.com/}.
For a fair comparison, we only compare with methods with released source codes.
We randomly select $20$ source images and $20$ reference images from the MT test set, MT-Wild dataset, and MT-HR test set.
After applying all makeup transfer methods, we obtain $400$ images for each method and each dataset respectively.
Five workers are asked to choose the best image among the six results according to the image realism and the similarity with reference makeup styles.
The human evaluation results are shown in Table \ref{t2}.
It is obvious that both  PSGAN  and PSGAN++ outperform other methods by a large margin.
By incorporating the newly proposed makeup detail loss, PSGAN++ can better transfer the makeup details and thus performs better than PSGAN in user studies.

\blue{\textbf{Expression/pose preservation}. Since our method only changes the color distributions, the expression and pose are naturally preserved between source and transferred images.
Since the expression and pose generally depend on the positions of facial landmarks, we need to validate the landmark preservation to prove the expression/pose preservation. Specifically, we calculate the cosine similarity $cos\_sim\in[0,1]$ between the detected landmarks of the source image $s$ and transferred image $t$ by:
\begin{equation}
    \begin{gathered}
      l_s = [x_1^s, y_1^s, x_2^s, y_2^s, \dots, x_n^s, y_n^s] = \operatorname{landmark\_detector}(s), \\
      l_t = [x_1^t, y_1^t, x_2^t, y_2^t, \dots, x_n^t, y_n^t] = \operatorname{landmark\_detector}(t), \\
      cos\_sim = \frac{l_s \cdot l_t}{\|l_s\|\|l_t\|}, 
   \end{gathered}
\end{equation}
where $n$ indicates the number of facial landmarks and is set to $68$ in our experiment, $x_i^s$ and $y_i^s$ indicate the coordinate of the $i$-th landmark of source image $s$, $x_i^t$ and $y_i^t$ indicate the coordinate of the $i$-th landmark of transferred image $t$.
The $68$ landmarks are detected by a pre-trained MTCNN \cite{zhang2016joint} model.
The $cos\_sim$ of our method and baselines are shown in Table \ref{tb:t1}. It shows that all the methods can well preserve the expression and pose.
}

\blue{\textbf{Identity preservation}. To validate the identity preservation when performing makeup transfer, we use ArcFace \cite{deng2019arcface} to calculate the similarity between the faces before and after makeup transfer. The results are shown in Table \ref{tb:t3}.
The average similarity of our method on MT dataset \cite{Li2018BeautyGANIF} is $0.9726$, which is slightly higher than other methods which also achieve a very high score.
The face verification result has demonstrated that our makeup transfer methods can preserve the identities well.
We also calculate the FID value between the reference and transferred image set, as shown in the second row in Table \ref{tb:t3}.
Our method also receives the best result among all the methods.}

\textbf{Qualitative comparison.} Figure \ref{res1} shows the qualitative comparison of PSGAN++ with other state-of-the-art methods on frontal faces in neutral expressions.
Since the code of BeautyGlow and PairedCycleGAN is not released, we crop the results from their corresponding papers.
The results produced by DIA have an unnatural color of hair and background since the transfer is performed over the whole image.
CycleGAN can only synthesize a general makeup style that is not similar to the reference.
Besides, BeautyGlow fails to preserve the color of pupils and does not have the same foundation makeup as reference.
We also use the pre-trained model released by the authors of LADN, which produces blurry transfer results and unnatural background.
PSGAN is able to generate vivid images with the same makeup styles as reference.
PSGAN++ can produce even better results.
For example, in the first row, PSGAN++ better preserves the structure of the lip.
In the second row, the foundation color of the PSGAN++ result is closer to the reference.

\begin{table}[t]
    \centering
    \blue{
    \setlength{\tabcolsep}{1.1mm}{
        \begin{tabular}{@{}ccccc@{}}
        \toprule
        Method   & PSGAN++ & PSGAN \cite{jiang2020psgan}  & BeautyGAN \cite{Li2018BeautyGANIF} & LADN \cite{Gu2019LADNLA} \\ \midrule
        cos\_sim $\uparrow$ & \textbf{0.9992}  & \textbf{0.9992} & \textbf{0.9992} &  0.9812 \\ \bottomrule
        \end{tabular}
    }
    \caption{Our method is expression/pose preserving by calculating the cosine similarity between landmarks.}
    \vspace{-4mm}
    \label{tb:t1}
    }
\end{table}

\begin{table}[t]
    \centering
    \setlength{\tabcolsep}{1.1mm}{
    \blue{
    \begin{tabular}{@{}ccccc@{}}
    \toprule
    Method   & PSGAN++ & PSGAN \cite{jiang2020psgan}  & BeautyGAN \cite{Li2018BeautyGANIF} & LADN \cite{Gu2019LADNLA} \\ \midrule
    ArcFace $\uparrow$    & \textbf{0.9726} & 0.9721 & 0.9691 & 0.8139 \\ \midrule
    FID  $\downarrow$   & \textbf{41.17}  & 41.98 & 44.61 & 64.25  \\ \bottomrule
    \end{tabular}}
     }
     \blue{
    \caption{Our method is identity preserving using ArcFace and FID metric.}
    \vspace{-8mm}
    \label{tb:t3}
    }
\end{table}

Figure \ref{wild} shows the qualitative comparison on images with different poses and expressions with the state-of-the-art methods (BeautyGAN and LADN) with available codes and pre-trained models.
Since traditional methods focus on frontal images, the results are unsatisfactory when dealing with images with different poses and expressions.
For example, the lip gloss is transferred to the skin in the first row of Figure \ref{wild}.
In the second row, other methods fail to perform the transfer on faces of different sizes.
Comparatively, our AMM module can accurately assign the makeup for every pixel by calculating the similarities, which makes our results look better.
While both PSGAN and PSGAN++ could handle different poses and expressions, PSGAN++ could generate even better results.
For instance, PSGAN (the 5th column) does not transfer the right eye shadow well.

\subsection{Ablation Studies}

\begin{figure}[!t]
    \includegraphics[width=1\linewidth]{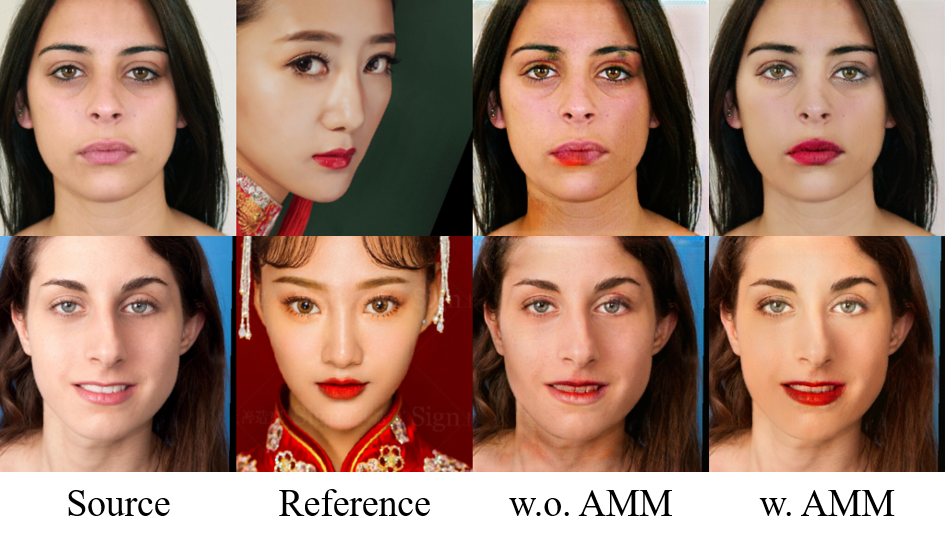}
    \caption{AMM module (the 4th column) could resolve the bad result (the 3rd column) due to pose and expression differences between source and reference images.}
    \label{ab1}
    \vspace{-4mm}
\end{figure}

\textbf{Attentive makeup morphing module.}
In PSGAN++, the AMM module is designed for alleviating the pose and expression differences between source and reference images, which morphs the distilled makeup matrices from $\gamma_{y_1}$ and $\beta_{y_1}$ to $\tilde{\gamma}_{y_1}$, $\tilde{\beta}_{y_1}$.
Figure \ref{ab1} shows the effectiveness of the AMM module.
The first row shows the case of different poses.
Without the AMM module, the foundation color of the transferred image is inconsistent due to the different pose from the reference.
By applying the AMM module, the foundation color is consistent and close to the reference.
The second row shows the case of different expressions.
The expression of the source image is smiling but the reference image has a neutral expression.
Without the AMM module, the lip gloss of the reference is applied on the tooth and part of the lip of the source, while using the AMM module could resolve this problem and correctly apply the lip gloss from the reference to the source.
The experiments above indicate that the AMM module could build the correspondence between a pixel from the source image and the corresponding one in the reference image rather than directly copying the makeup from the same absolute location.

\begin{figure}[!t]
    \includegraphics[width=1\linewidth]{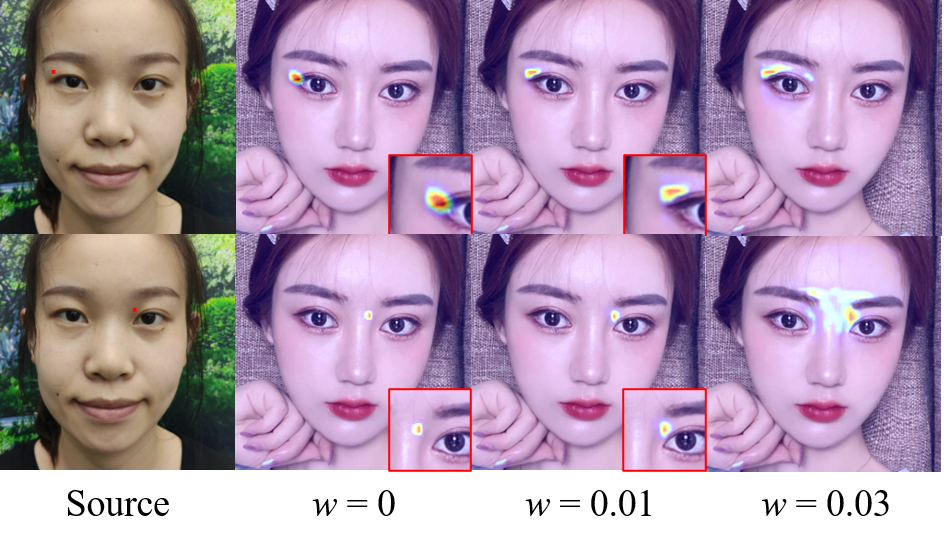}
    \caption{The attention maps with different weights on visual features given a red point on the skin in the source image (the 1st column).}
    \label{ab2}
    \vspace{-4mm}
\end{figure}

\begin{figure*}[!t]
    \includegraphics[width=1\linewidth]{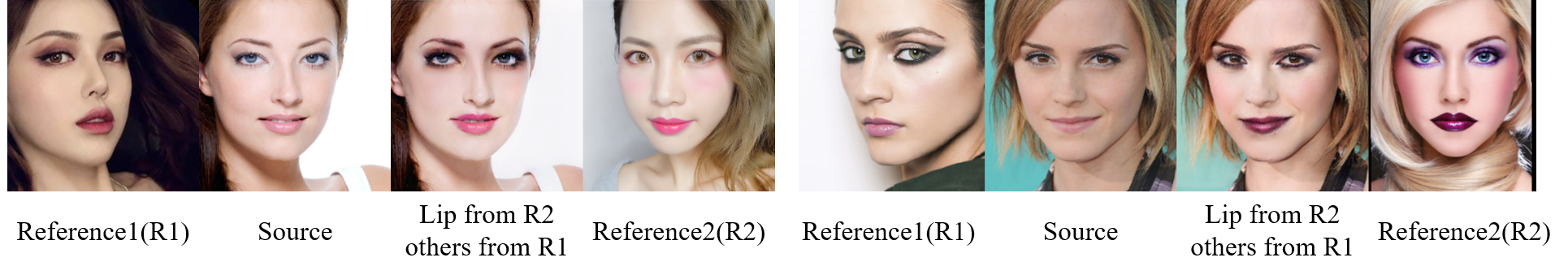}
    \caption{Partial makeup transfer results. Given a source image, the transferred images are generated by transferring the lipstick from the reference 2 and other makeup from the reference 1. }
    \label{partial}
    \vspace{-3mm}
\end{figure*}

\begin{figure*}[!t]
    \includegraphics[width=1\linewidth]{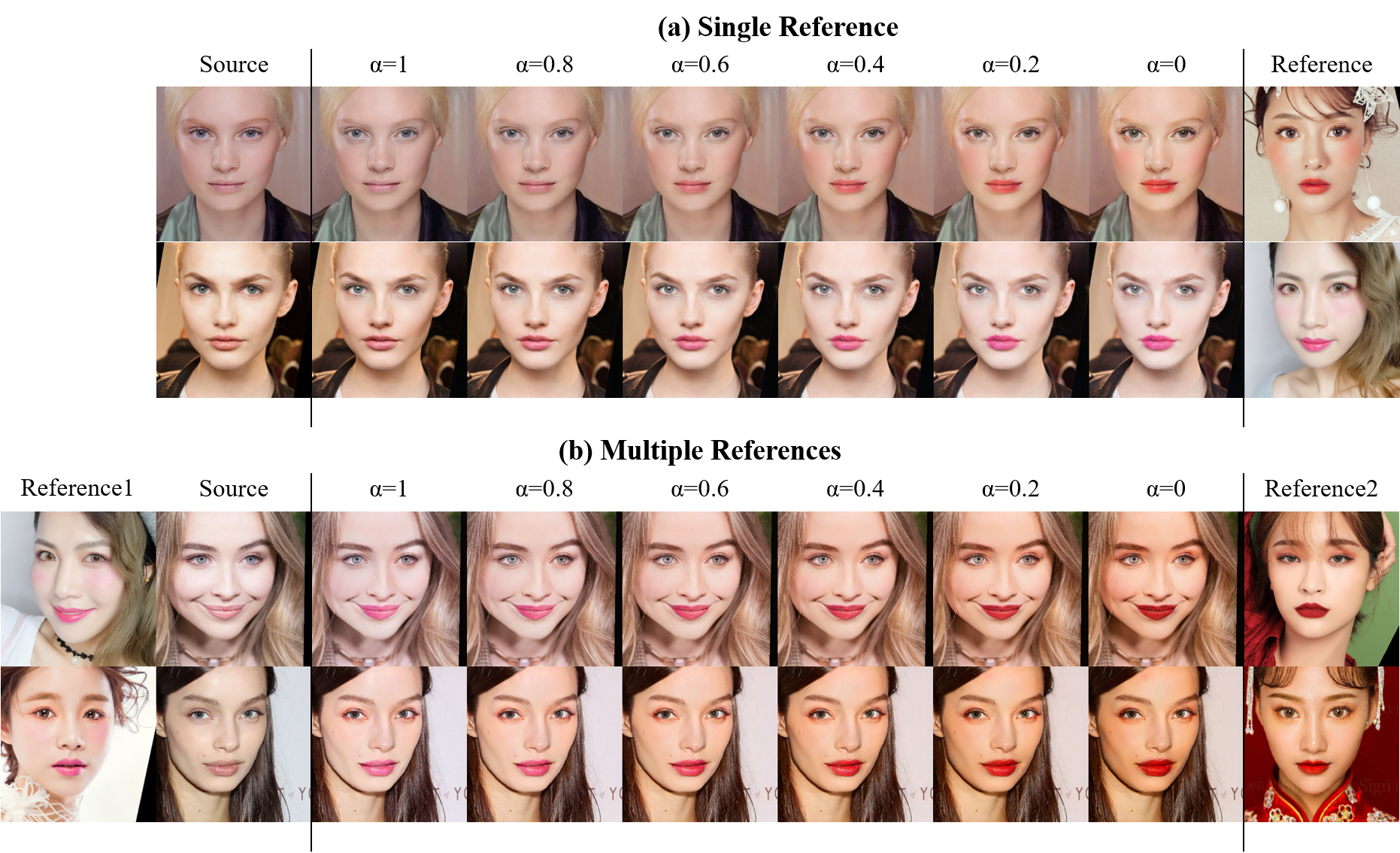}
    \caption{Interpolated makeup transfer results. The 1st row uses one reference to adjust the degree of makeup. The 2nd row uses two references to gradually change the source from reference 1 (left) towards reference 2 (right).}
    \label{interpolated}
    \vspace{-4mm}
\end{figure*}

\textbf{The weight of visual feature in calculating $A$.}
The attentive matrix $A$ in the AMM module is calculated by considering both the visual features $\mathbf{v}$ and the relative positions $\mathbf{p}$ using Eq. (\ref{equ5}).
Figure \ref{ab2} demonstrates the ablation study of the visual feature weight.
Without using the visual feature, the attention maps are similar to a 2D Gaussian distribution (the 2nd column).
In this case, the attention maps could not reflect the correct regions on the face well.
For instance, in the first row, the attention map crosses the boundary of the eye.
Similarly, the attention map fails to avoid the bridge of the nose in the second row.
Besides, a larger visual weight might make the attention maps scattered and unreasonable (the 4th column).
We empirically set the visual weight to $0.01$ and the results (the 3rd column) show that the attention maps could focus more on the skin and also bypass the eyes as well as the bridges of the nose.

\begin{figure*}[t]
    \includegraphics[width=1\linewidth]{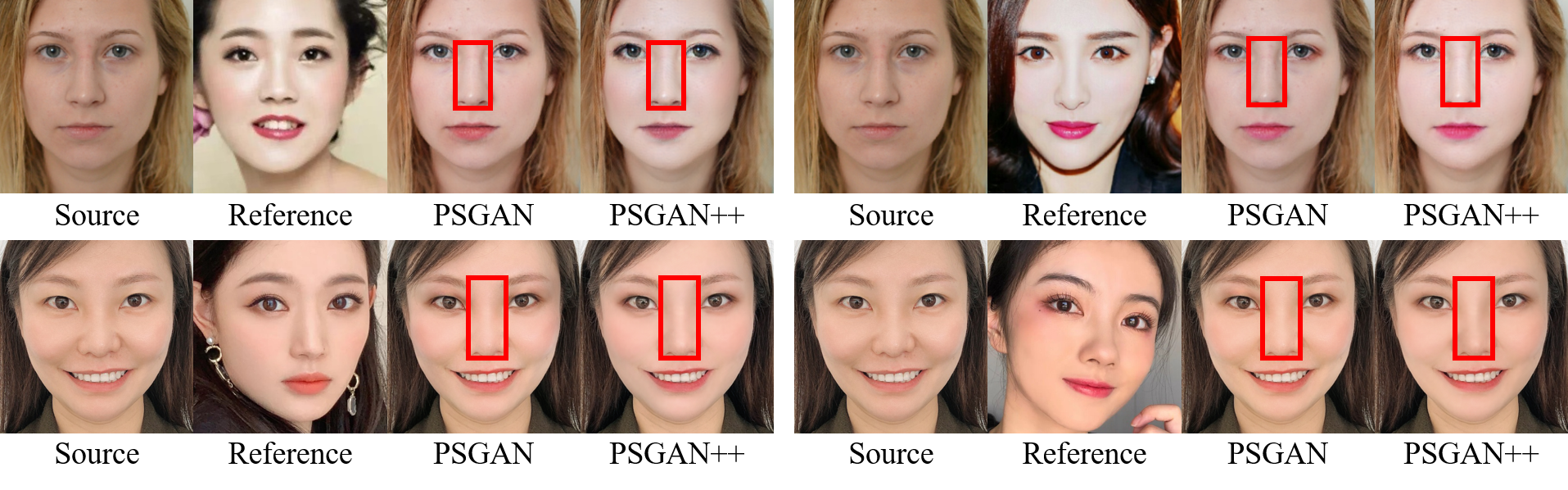}
        \vspace{-7mm}
    \caption{The comparison between PSGAN and PSGAN++ when transferring the
        \emph{highlights} on MT dataset (first row) and MT-HR dataset (second row). Best viewed in color.}
    \label{detailed_resutls_highlight}
    \vspace{-2mm}
\end{figure*}

\begin{figure*}[t]
    \includegraphics[width=1\linewidth]{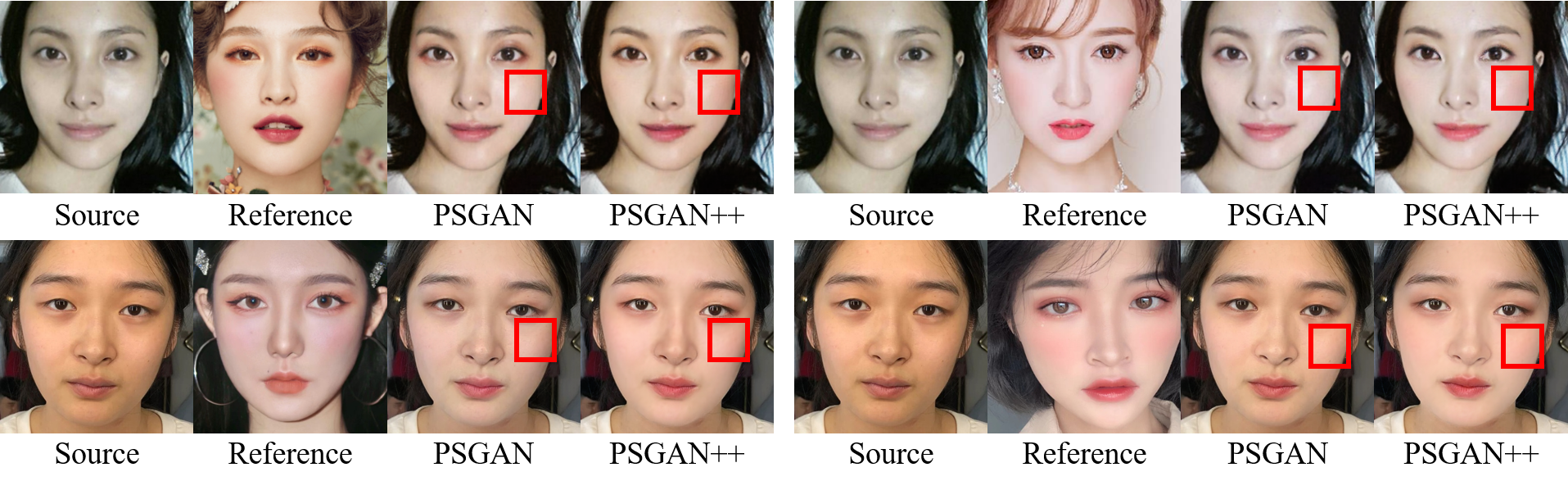}
        \vspace{-7mm}
    \caption{The comparison between PSGAN and PSGAN++ when transferring the \emph{blush} on MT dataset (first row) and MT-HR dataset (second row). Best viewed in color.}
    \label{detailed_resutls_blush}
        \vspace{-1mm}
\end{figure*}

\begin{figure*}[!t]
    \centering
    \includegraphics[width=0.7\linewidth]{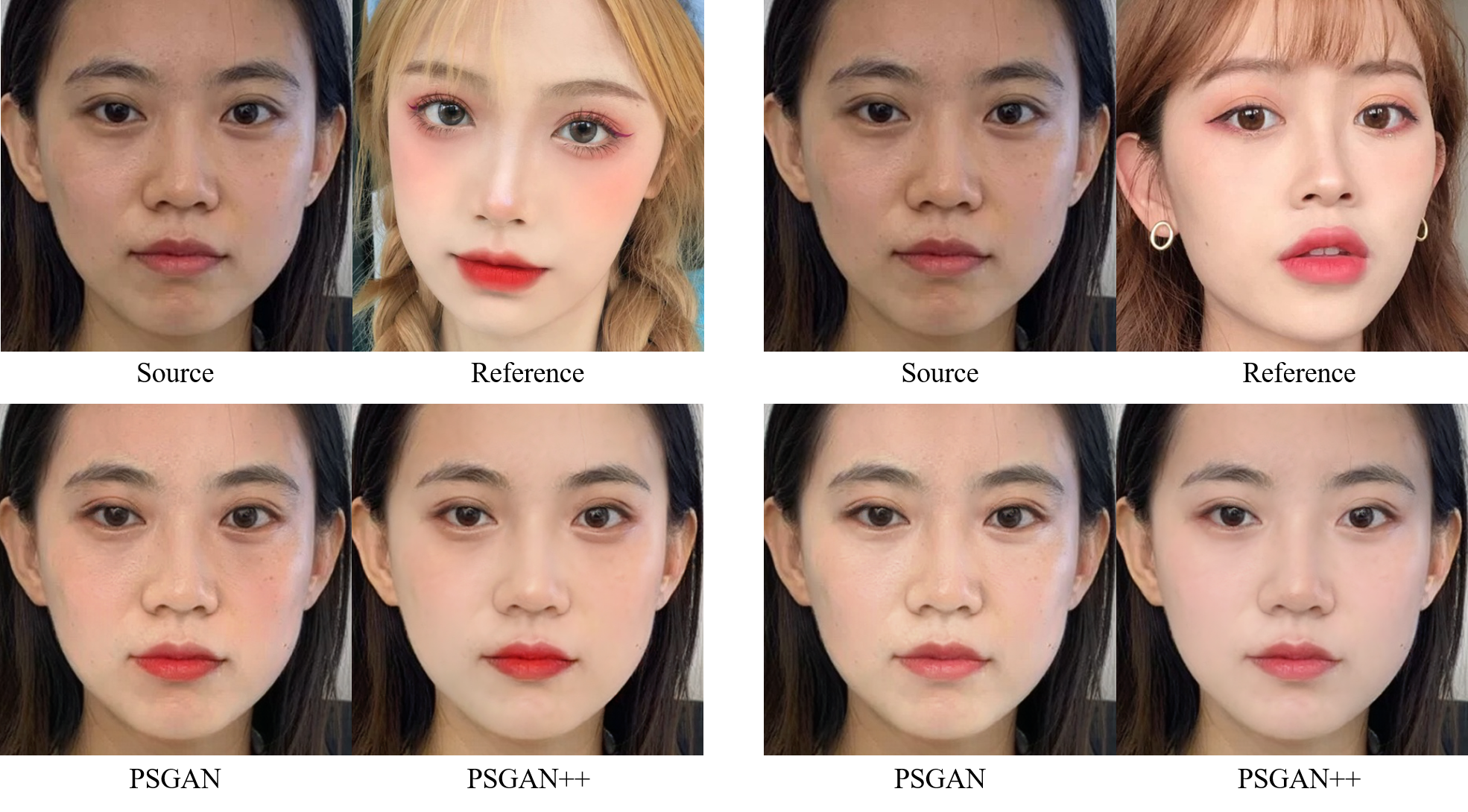}
        \vspace{-2mm}
    \caption{The comparison between PSGAN and PSGAN++ when handling the \emph{high resolution} images.}
    \label{high_resolution}
        \vspace{-3mm}
\end{figure*}

\subsection{Characteristics of makeup Transfer}

\begin{figure*}[!t]
    \includegraphics[width=1\linewidth]{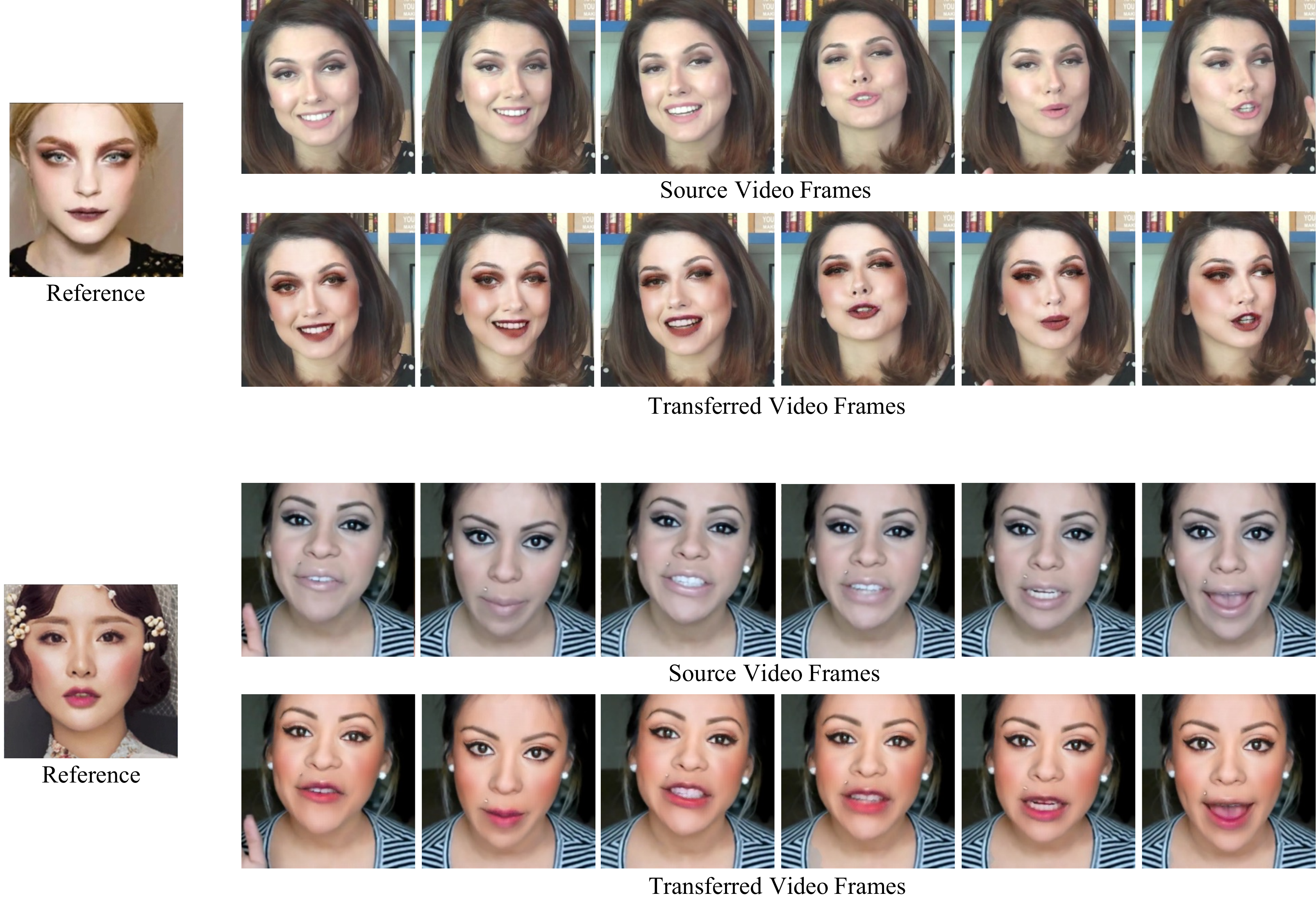}
    \vspace{-2mm}
    \blue{
    \caption{Our model  produces good and stable video makeup transfer results. }
    \vspace{-3mm}
    \label{video}
    }
 \end{figure*}

\textbf{Partial makeup transfer.}
Due to the spatial-aware characteristic of the makeup tensors, PSGAN++ could achieve partial and interpolated makeup transfer during testing.
Partial makeup generation could be realized by computing new makeup tensors, which are weighted by face parsing results.
Let $x$, $y_1$, and $y_2$ denote the source image and two reference images.
$\mathbf{\tilde{\Gamma}_x}$, $\mathbf{\tilde{B}_x}$ and $\mathbf{\tilde{\Gamma}_{y_1}}$, $\mathbf{\tilde{B}_{y_1}}$ as well as $\mathbf{\tilde{\Gamma}_{y_2}}$, $\mathbf{\tilde{B}_{y_2}}$ could be obtained by feeding the images to MDNet.
Besides, the face parsing mask $m_x$ of $x$ could be obtained through the method in \cite{Zhao2016PyramidSP}.
For example, if we want the lipstick form $y_1$ and other makeup from $y_2$, we could first obtain the binary mask of the lip $m^{l}_x \in \{0, 1\}^{H \times W}$.
In this way, PSGAN++ could achieve partial makeup transfer by applying different makeup tensors on corresponding areas.
Based on Eq. (\ref{equ1}), the partial transferred feature map $\mathbf{\tilde{V}_x}$ is calculated by
    \begin{equation}
        \small
        \mathbf{\tilde{V}_x} = (m^{l}_x \mathbf{\tilde{\Gamma}_{y_1}} + (1-m^{l}_x)\mathbf{\tilde{\Gamma}_{y_2}}) \mathbf{V_x}  + (m^{l}_x \mathbf{\tilde{B}_{y_1}}  + (1-m^{l}_x) \mathbf{\tilde{B}_{y_2}}).
        \label{equ21}
        \end{equation}
Figure \ref{partial} shows the partial makeup transfer results from two references. The third columns show the results of recombining the makeup of the lip from reference 2 and other parts of makeup from reference 1, which looks natural and realistic. Also, by assigning $y_2 = x$, PSGAN++ could only transfer the lipstick from reference 1 and remain other parts unchanged. The partial makeup transfer makes PSGAN++ more flexible and practical.

\textbf{Interpolated makeup transfer.} Furthermore, we could interpolate the makeup from two reference images by a coefficient $\alpha \in [0,1]$.
By weighting the makeup tensors from two references $y_1$ and $y_2$ with the coefficient $\alpha$, new makeup tensors could be obtained.
The interpolated feature map $\mathbf{\tilde{V}_x}$ is calculated by
\begin{equation}
\small
\mathbf{\tilde{V}_x} = (\alpha \mathbf{\tilde{\Gamma}_{y_1}} + (1-\alpha)\mathbf{\tilde{\Gamma}_{y_2}}) \mathbf{V_x}  + (\alpha \mathbf{\tilde{B}_{y_1}} + (1-\alpha) \mathbf{\tilde{B}_{y_2}}).
\label{equ3}
\end{equation}
The results of interpolated makeup transfer with both one and two reference images are shown in Figure \ref{interpolated}. By adjusting the coefficient $\alpha$, we could generate a series of images transiting from reference 1 to reference 2. Similarly, by assigning $y_1 = x$, we could adjust the degree of transfer using only one reference. The experimental results indicate that PSGAN++ could realize degree-controllable makeup transfer by using one or two references.

Thanks to the spatial-aware makeup tensors, we could also realize partial and interpolated makeup transfer simultaneously. The aforementioned experiments also demonstrate that PSGAN++ has significantly broadened the application of makeup transfer.

\textbf{Detail-preserving makeup transfer.}
The comparison between PSGAN and PSGAN++ when transferring the highlight is shown in Figure \ref{detailed_resutls_highlight}.
When applying the models to $512 \times 512$ images, we add an additional layer in both the encoder and decoder.
By incorporating the makeup detail loss, PSGAN++ is more successful when transferring the highlight on references' noses compared to PSGAN.
Figure \ref{detailed_resutls_blush} shows the comparison between PSGAN and PSGAN++ when transferring the blush on references. Unlike lipstick and eye shadow which are usually evident on faces, blush is not obvious and is more difficult to transfer from references. Regardless of this difficulty, PSAGN++ could still learn the blush on the reference successfully, while PSGAN could not transfer this kind of detailed information well.
By incorporating the makeup detail loss on the selected area, PSGAN++ has a stronger ability to transfer makeup details (e.g. highlight, blush) from references.

\textbf{High-resolution makeup transfer.}
The comparison between PSGAN and PSGAN++ is shown in Figure \ref{high_resolution}.  For both PSGAN and PSGAN++, we feed the $512 \times 512$ images from the MT-HR dataset into the framework. Due to the superiority of makeup detail loss, PSGAN++ is able to produce better and clearer results, especially in the areas of makeup details like cheek and nose.

\vspace{-2mm}
\subsection{Video Makeup Transfer}
Compared with image makeup transfer, video makeup transfer is a more challenging but meaningful task with wide and prosperous applications. The pose and expression from each frame continuously change in the video, which brings an extra challenge. To examine the effectiveness of our network, we apply makeup transfer on each frame of the video, as shown in Figure \ref{video}. Although PSGAN++ is not intentionally designed for video makeup transfer, it produces good and stable visual results. More results about video makeup transfer could be found in our \emph{supplementary materials}.
\blue{In some cases, the makeup transfer may change the background slightly. The reason is that the adversarial loss is applied to the whole image, thus the hue of the background may change a little to be compatible with the foreground. By using the face parsing maps of the video frames, the background can stay unchanged by blending the background of the source image with the foreground of the generated image.}

\subsection{Makeup Removal}

Besides makeup transfer, PSGAN++ could also remove the makeup on faces. A good makeup removal algorithm has two requirements. First, it can visually remove the makeup. Second, it will not change the identity information after removing makeup.

\begin{table}[!t]
    \centering
    \setlength{\tabcolsep}{1mm}{
        \begin{tabular}{@{}ccccccc@{}}
            \toprule
            Method   & CycleGAN & BLAN & BLAN-2 & DRL &PSGAN++ \\ \midrule
            Rank-1    & 88.65 & 91.03 & 91.21 &93.08 &  \textbf{93.27}   \\ \midrule
            TPR@FPR =0.1\%   &57.00  & 55.62  & 57.08 &71.07 &   \textbf{77.44}    \\ \midrule
            TPR@FPR =1\% & 82.78 & 85.29 & 83.94 &91.27 & \textbf{92.92} \\
            \bottomrule
    \end{tabular}}
    \caption{The face verification results on the CMF dataset (\%).}
    \label{verification}
    \vspace{-8mm}
\end{table}

\begin{figure}[!t]
        \includegraphics[width=1\linewidth]{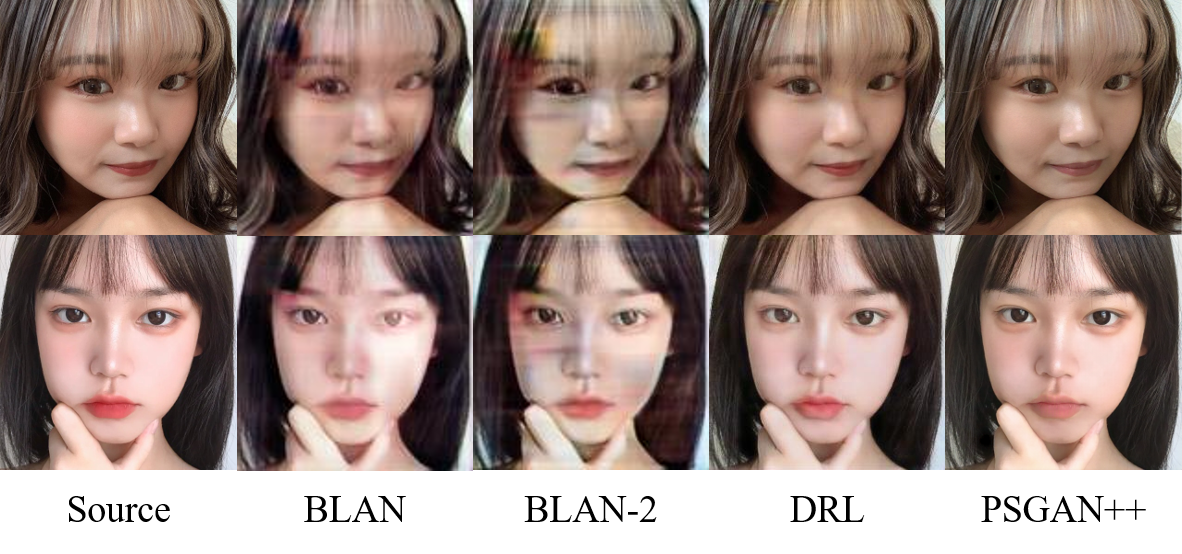}
        \vspace{-7mm}
        \caption{The visualization results of makeup removal.}
        \label{demakeup}
        \vspace{-2mm}
\end{figure}
We compare our result with BLAN \cite{li2017anti}, BLAN2 \cite{li2019learning} and DRL \cite{li2019disentangled}, as shown in Figure \ref{demakeup}, and find that our visual result has successfully removed the makeup (e.g. lipstick) on the source image. But BLAN and BLAN-2 suffer from unclear visual results and DRL does not remove the lipstick well.

For the second aspect, we conduct face verification between de-makeup images and the original non-makeup images on CMF dataset \cite{li2019disentangled}, as shown in Table \ref{verification}.
\emph{Cross-Makeup Face} (CMF) dataset contains $2,611$ pairs of $1,425$ identities. Compared with most existing makeup datasets, the CMF dataset contains identity information, which enables face verification experiments. We follow the experimental setting in \cite{li2019disentangled}. In concrete, the feature extractor we use is a pre-trained LightCNN-9 \cite{2018A} model and we adopt cosine distance to measure the similarity between features. Compared with CycleGAN \cite{Zhu2017UnpairedIT}, BLAN \cite{li2017anti}, BLAN2 \cite{li2019learning} and DRL \cite{li2019disentangled}, PSGAN++ achieves the best performance on three metrics, indicating that our method preserves the features of human faces well and successfully removes the makeup on faces.

\begin{figure}[!t]
    \centering
    \includegraphics[width=0.8\linewidth]{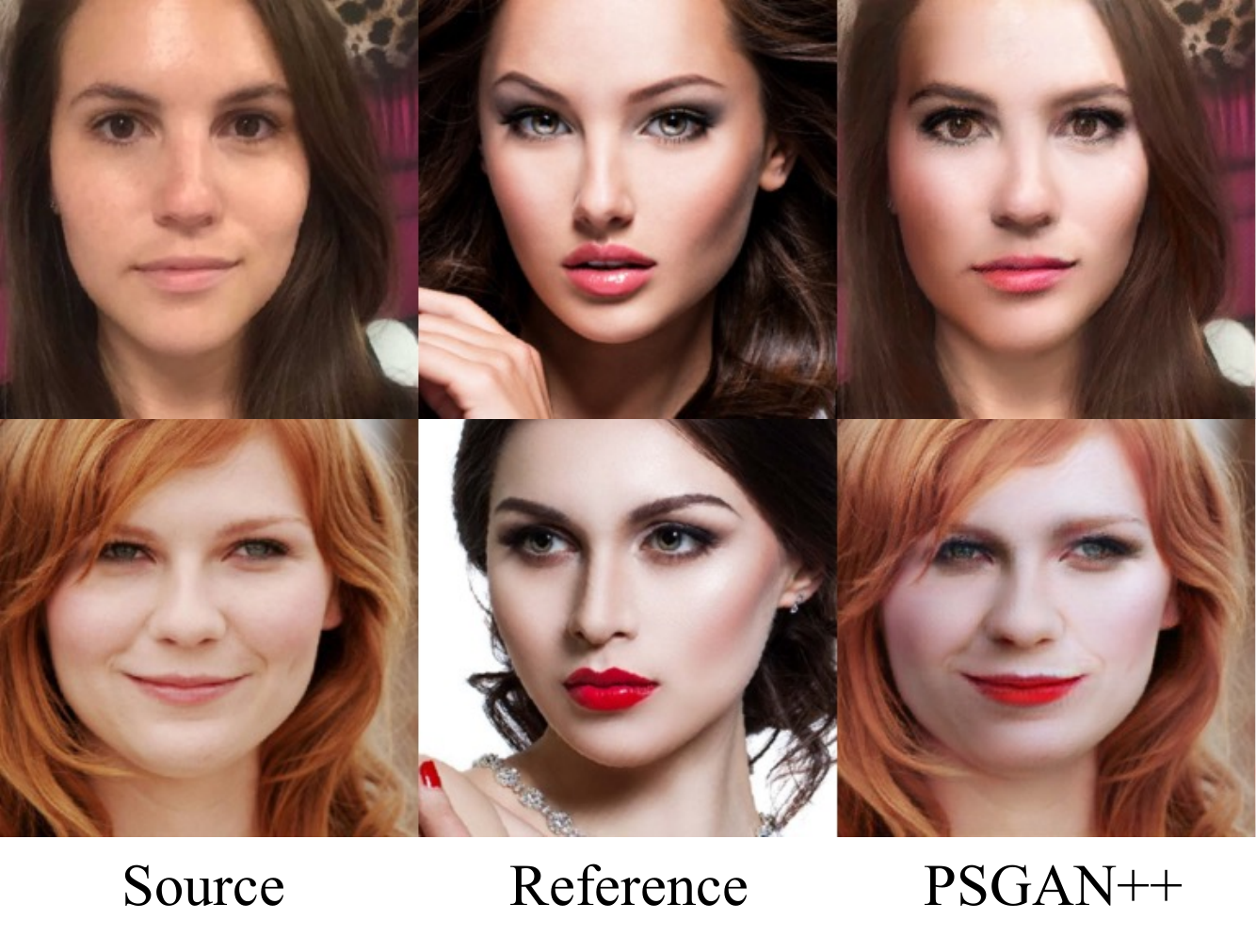}
    \blue{
    \caption{Failure cases.}
    \label{fail}
    }
    \vspace{-3mm}
\end{figure}

\subsection{\blue{Limitations}}
\blue{
PSGAN++ can be further improved in terms of the following two aspects. 
First, our method may wrongly transfer the shadow from the reference image.
As shown in  Figure \ref{fail}, the illumination of source images and reference images are very different. The source images (the 1st column) are clear and are illuminated by the light from the front, while the reference images (the 2nd column) are illuminated by the light from the side and have shadows on the skin. Our method then transfers the color of shadows and obtains abnormal results.
Second, we have studied the $512\times512$ makeup transfer problem via the newly collected MT-HR dataset, which is already much larger than existing makeup transfer datasets. However, in some real applications, facial images are with even higher resolution, such as $1024\times1024$.
}

\section{Conclusion and \blue{Future Work}}
In this paper, we propose the PSGAN++ aiming to enable makeup transfer and removal in real-world applications.
For makeup transfer, PSGAN++ first distills the makeup information from the reference image into makeup matrices using MDNet. Then, PSGAN++ leverages a newly proposed AMM module for accurate transfer between different identities. Besides, a makeup detail loss is proposed to provide pixel-level supervision to achieve highlight and blush makeup transfer.
For makeup removal, PSGAN++ distills the identity information from the source image into identity matrices using IDNet.
The STNet leverages the obtained makeup matrices and identity matrices to achieve makeup transfer and removal respectively by scaling or shifting the visual feature map for only once.
In general, PSGAN++ can achieve pose and expression robust, partial, degree-controllable, and detail-preserving makeup transfer, which greatly broadens the application range of makeup transfer.
Extensive experiments on three datasets demonstrate our approach can achieve state-of-the-art transfer results on both frontal facial images and facial images that have various poses and expressions.
Both the code and the collected datasets will be released.
Moreover, we plan to apply our novel framework to other conditional image synthesis problems that require customizable and precise synthesis.

\blue{
    As for future directions, we plan to explore the 3D face modeling method to eliminate the interference of illumination.
    Besides, we plan to study the makeup transfer of higher resolution images, which can further bring makeup transfer technology to real-world applications.
}

\section{Acknowledgements}
We sincerely thank Jiayun Wang for his efforts and Wenyan Wu, Chengyao Zheng for their useful comments.
This research is supported in part by National Natural Science Foundation of China (Grant 61876177), Beijing Natural Science Foundation (4202034, Grant No. JQ18017), Zhejiang Lab (No. 2019KD0AB04).

\bibliographystyle{plain}
\bibliography{egbib}

\begin{IEEEbiography}[{\includegraphics[width=1in,height=1.25in,clip,keepaspectratio]{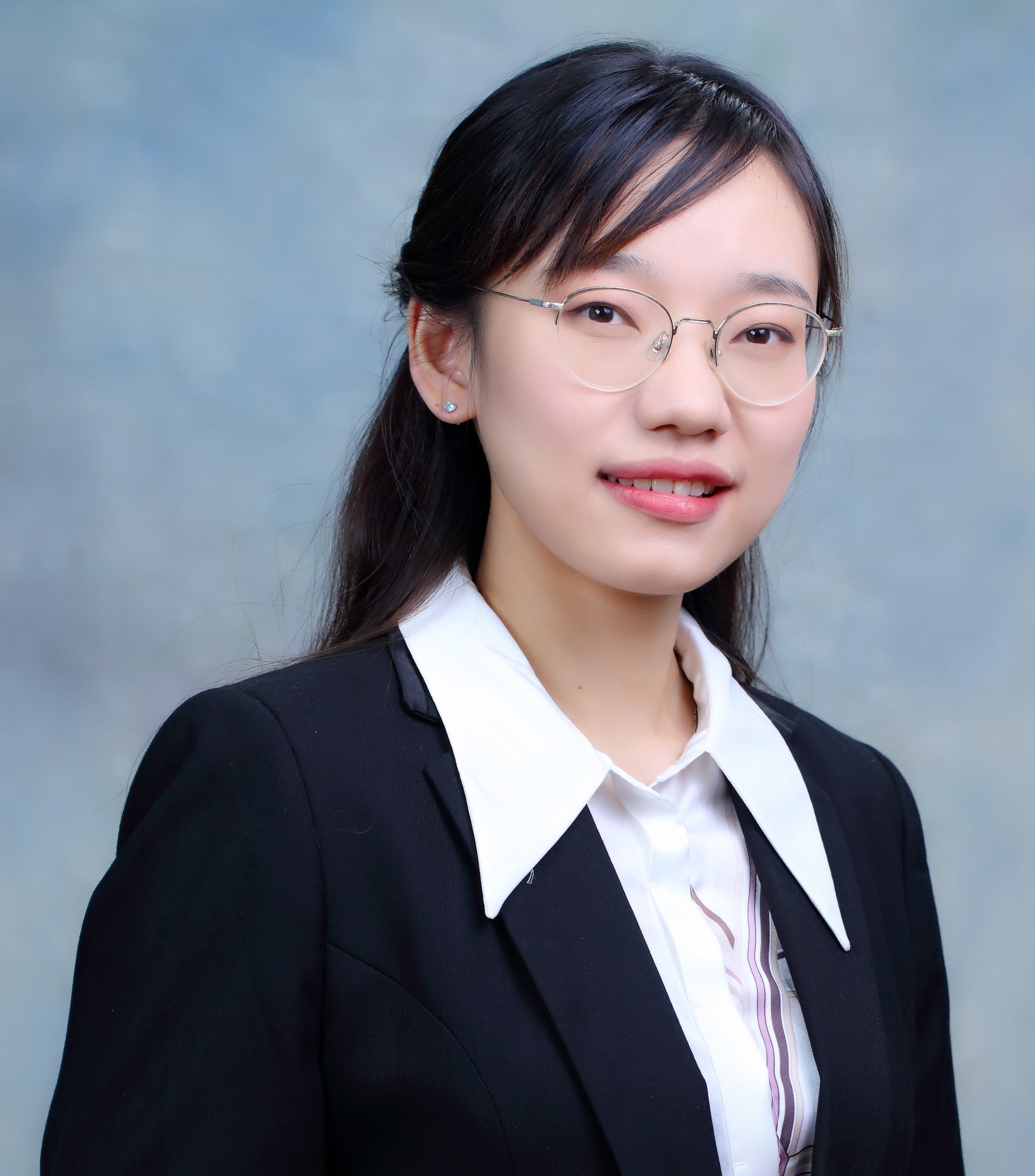}}]{Si Liu} is currently an associate professor in Beihang University. She received her Ph.D. degree from Institute of Automation, Chinese Academy of Sciences. She has been Research Assistant and Postdoc in National University of Singapore. Her research interest includes computer vision and multimedia analysis. She has published over 40 cutting-edge papers on the human-related analysis including the human parsing, face editing and image retrieval. She was the recipient of Best Paper of ACM MM 2013, Best demo award of ACM MM 2012. She was the Champion of CVPR 2017 Look Into Person Challenge and the organizer of the ECCV 2018 Person in Context Challenge.
\end{IEEEbiography}

\begin{IEEEbiography}[{\includegraphics[width=1in,height=1.25in,clip,keepaspectratio]{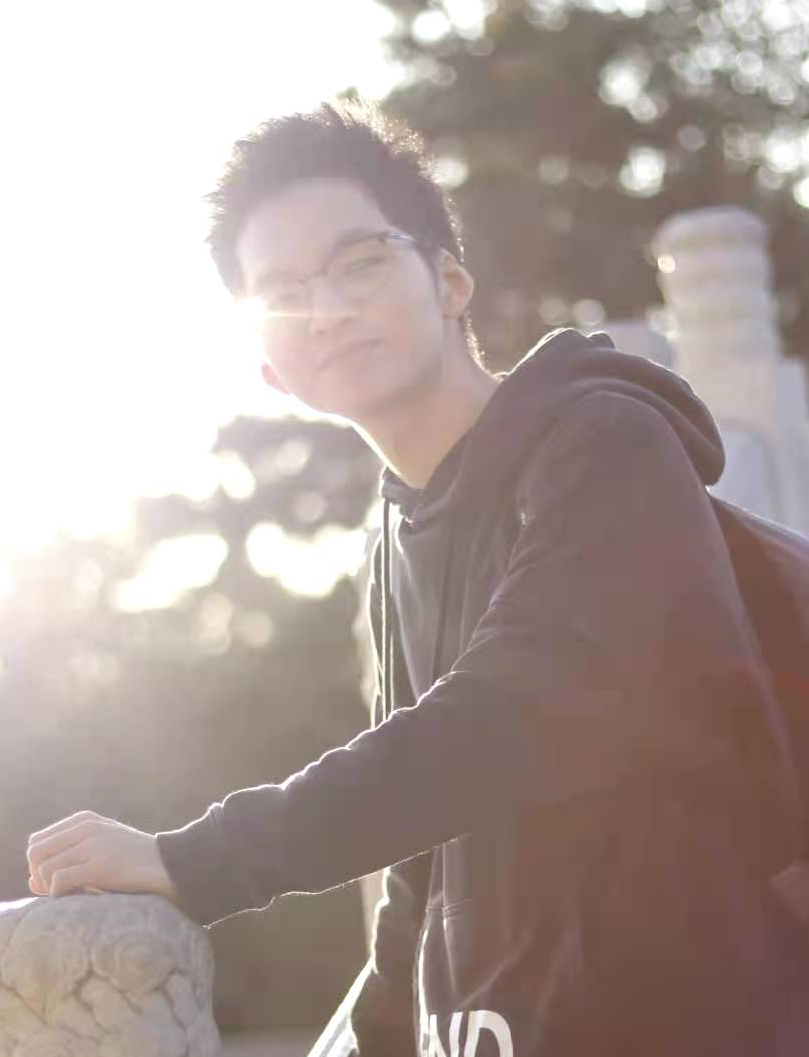}}]{Wentao Jiang} is currently a Ph.D. student at School of Computer Science and Engineering, Beihang University (BUAA). Before that, he was a master student at Beihang University. He received his B.Eng. degree from Harbin Engineering University (HEU) in 2019. His research interests include image generation and multimedia.
\end{IEEEbiography}

\begin{IEEEbiography}[{\includegraphics[width=1in,height=1.25in,clip,keepaspectratio]{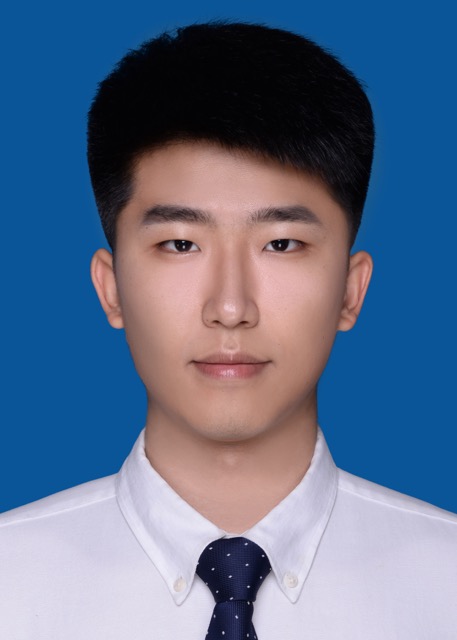}}]{Chen Gao} is currently a Ph.D. student at Beihang University. He received the master’s degree from the Institute of Information Engineering, Chinese Academy of Sciences and the B.S. degree from Xidian University. His research interests include object detection, image generation, multimedia and AutoML.
\end{IEEEbiography}

\begin{IEEEbiography}[{\includegraphics[width=1in,height=1.25in,clip,keepaspectratio]{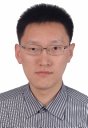}}]{Ran He}  received the BE degree in computer science from the Dalian University of Technology, the MS degree in computer science from the Dalian University of Technology, and the PhD
    degree in pattern recognition and intelligent systems from the Institute of Automation, Chinese Academy of Sciences in 2001, 2004 and 2009, respectively. Since September 2010, he has
    joined NLPR where he is currently professor. He currently serves as an associate editor of Neurocomputing (Elsevier) and serves on the program committee of several conferences. His research interests focus on information theoretic learning, pattern recognition, and computer vision. He is a senior member of IEEE and a fellow of IAPR.
\end{IEEEbiography}

\begin{IEEEbiography}[{\includegraphics[width=1in,height=1.25in,clip,keepaspectratio]{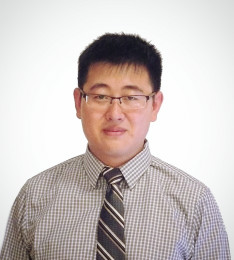}}]{Jiashi Feng} is currently an assistant professor with the Department of Electrical and Computer Engineering at National University of Singapore. His research areas include machine learning and their applications in computer vision and AI. He has authored/co-authored more than 100 technical papers on deep learning, robust machine learning, image classification, object detection, face recognition. He received the best technical demo award from ACM MM 2012, best paper award from TASK-CV ICCV 2015, best student paper award from ACM MM 2018. He is also the recipient of Innovators Under 35 Asia, MIT Technology Review 2018. He served as the area chairs for ACM MM 2017, 2018 and program chair for ICMR 2017.
\end{IEEEbiography}

\begin{IEEEbiography}[{\includegraphics[width=1in,height=1.25in,clip,keepaspectratio]{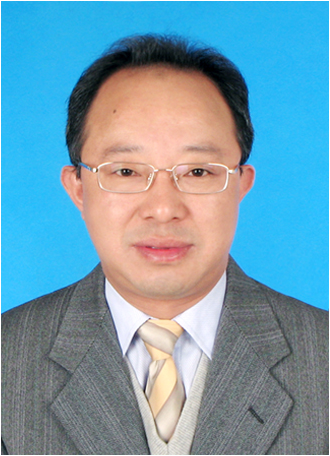}}]{Bo Li} is currently a Changjiang Distinguished Professor of School of Computer Science and Engineering, Beihang University. He is a recipient of The National Science Fund for Distinguished Young Scholars. He is currently the dean of AI Research Institute, Beihang University. He is the chief scientist of National 973 Program and the principal investigator of the National Key Research and Development Program. He has published over 100 papers in top journals and conferences and held over 50 domestic and foreign patents.
\end{IEEEbiography}

\begin{IEEEbiography}[{\includegraphics[width=1in,height=1.25in,clip,keepaspectratio]{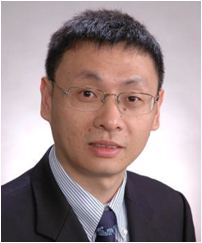}}] {Shuicheng Yan} is currently the director of Sea AI Lab (SAIL) and group chief scientist of Sea. He is an Fellow of Academy of Engineering, Singapore, IEEE Fellow, ACM Fellow, IAPR Fellow. His research areas include computer vision, machine learning and multimedia analysis. Till now, he has published over 600 papers in top international journals and conferences, with Google Scholar Citation over 40,000 times and H-index 105. He had been among “Thomson Reuters Highly Cited Researchers” in 2014, 2015, 2016, 2018, 2019. Dr. Yan’s team has received winner or honorable-mention prizes for 10 times of two core competitions, Pascal VOC and ImageNet (ILSVRC), which are deemed as “World Cup” in the computer vision community. Also his team won over 10 best paper or best student paper prizes and especially, a grand slam in ACM MM, the top conference in multimedia, including Best Paper Award, Best Student Paper Award and Best Demo Award.

\end{IEEEbiography}

\end{document}